%% file: main.tex
\documentclass[11pt]{article}

\input{macros}

\usepackage{hunyuan}

\usepackage[capitalize,noabbrev]{cleveref}
\usepackage{tabularx}
\usepackage[most]{tcolorbox}
\usepackage{wrapfig}
\usepackage{xspace}
\usepackage{fancyhdr}
\usepackage{bm}
\usepackage{placeins}
\usepackage{colortbl}

\makeatletter
\long\def\@makecaption#1#2{%
  \vskip 10pt
  \setbox\@tempboxa\hbox{#1: #2}%
  \ifdim \wd\@tempboxa >\hsize
    \noindent #1: #2\par   
  \else
    \hbox to\hsize{\hfil\box\@tempboxa\hfil}
  \fi}
\makeatother

\hypersetup{
  colorlinks=true,
  linkcolor=hunyuanblue,
  citecolor=hunyuanblue,
  urlcolor=hunyuanblue,
}

\makeatletter
\def\section{\@startsiction{section}{1}{\z@}{-0.24in}{0.10in}
             {\large\bf\raggedright\color{hunyuanblue}}}
\def\subsection{\@startsection{subsection}{2}{\z@}{-0.20in}{0.08in}
                {\normalsize\bf\raggedright\color{hunyuanblue}}}
\makeatother

\definecolor{abstractbg}{HTML}{F0F7FC}
\definecolor{caseblue}{RGB}{42, 91, 160}
\definecolor{casebg}{RGB}{246, 248, 252}
\definecolor{caseborder}{RGB}{180, 195, 220}
\definecolor{paperblue}{HTML}{1F77B4}
\definecolor{paperred}{HTML}{D62728}
\definecolor{deepred}{HTML}{B22222}
\definecolor{softred}{HTML}{C44E52}

\fancypagestyle{plain}{%
  \fancyhf{}%
  \fancyfoot[C]{\thepage}%
}

\fancypagestyle{firststyle}{%
  \fancyhf{}%
  \fancyhead[L]{\raisebox{-18.5mm}[0pt][0pt]{\includegraphics[height=12mm]{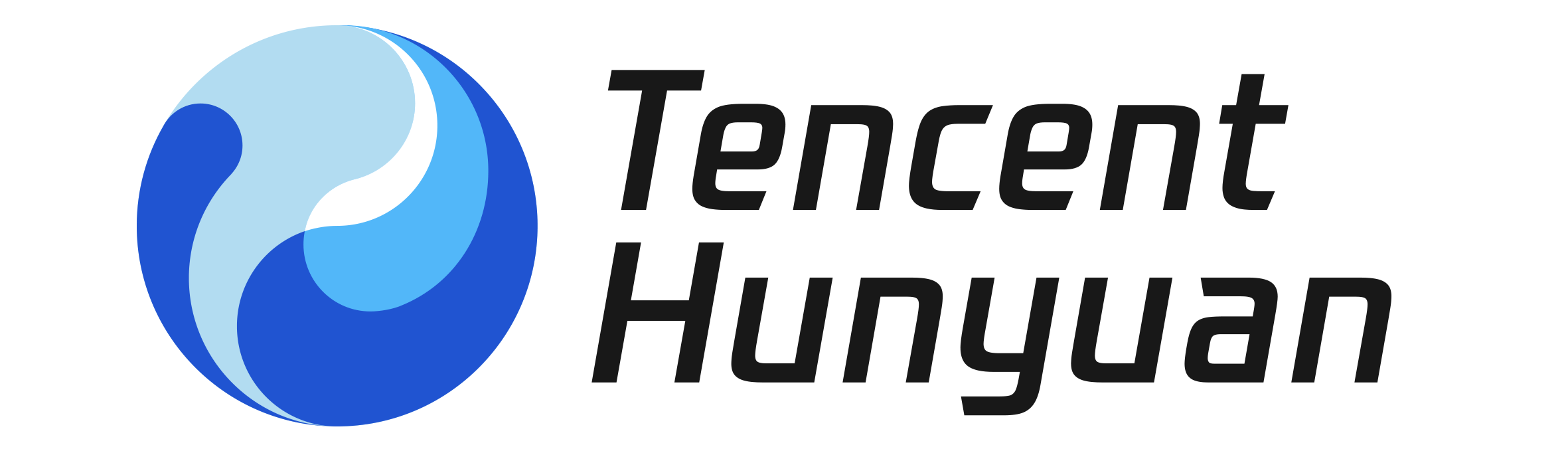}}}%
  \fancyfoot[C]{\thepage}%
}

\newcolumntype{Y}{>{\raggedright\arraybackslash}X}

\newtcolorbox[auto counter, number within=section]{compactcase}[2][]{
  breakable,
  enhanced,
  colback=gray!2,
  colframe=gray!30,
  colbacktitle=gray!12,
  coltitle=black,
  fonttitle=\bfseries,
  title={#2},
  boxrule=0.45pt,
  arc=1mm,
  left=1.5mm,
  right=1.5mm,
  top=1mm,
  bottom=1mm,
  toptitle=0.7mm,
  bottomtitle=0.7mm,
  before skip=0.8em,
  after skip=0.8em,
  label={#1}
}

\input{_define}

\newcommand{\papertitle}{Exploring the Design Space of\\Reward Backpropagation for Flow Matching}
\newcommand{\paperauthors}{%
\textbf{Ruoyu Wang}$^{1,2,*}$ \quad
\textbf{Boye Niu}$^{2,3,*}$ \quad
\textbf{Xiangxin Zhou}$^{2,*\,\mathparagraph\,\ddagger}$ \\
\textbf{Yushi Huang}$^{2,4}$ \quad
\textbf{Tongliang Liu}$^{3}$
\quad
\textbf{Chi Zhang}$^{1,\ddagger}$ \quad
}
\newcommand{\paperaffiliations}{%
$^1$Westlake University \quad $^2$Tencent Hunyuan \\[2pt]
$^3$University of Sydney \quad $^4$The Hong Kong University of Science and Technology \\[6pt]
{\small $^*$Equal contribution \quad $^\mathparagraph$Project Lead \quad $^\ddagger$Corresponding author}
}

\begin{document}

\thispagestyle{firststyle}
\vspace*{0.25cm}
{\color{hunyuanblue}\hrule height 0.6pt}
\vskip 6mm
\begin{center}
{\LARGE\bfseries \papertitle\par}
\end{center}
\vskip 3mm
{\color{hunyuanblue}\hrule height 0.6pt}
\vskip 6mm
\begin{center}
\paperauthors\\[4pt]
{\small \paperaffiliations}
\end{center}
\vskip 6mm

\begin{tcolorbox}[
  colframe=abstractbg,
  colback=abstractbg,
  boxrule=0pt,
  arc=2mm,
  enhanced,
  top=12pt,
  bottom=12pt,
  left=15pt,
  right=15pt,
  width=\textwidth,
]
\textbf{Abstract.}\quad
\input{sec/00_abs}

\vskip 8pt
\textbf{Date:} \today
\end{tcolorbox}

\input{sec/01_intro}
\input{sec/06_related}
\input{sec/02_background}
\input{fig/Flux1_overview_radar}
\input{sec/03_method}
\input{sec/04_experiment}
\input{sec/05_conclusion}

\subsubsection*{Acknowledgments}
We thank Zhanhao Liang and Yuchi Liu for insightful discussions and help.

\bibliography{reference}

\clearpage
\appendix

\input{A-sec/01_derivations}

\input{A-sec/02_experimental_details}
\input{A-sec/03_algorithm_templates}

\input{A-sec/03_qualitative_results}

\end{document}

%% file: macros.tex
\usepackage{amssymb,amsmath,amsthm,bbm}
\usepackage{verbatim,float,url,dsfont}
\usepackage{graphicx,subcaption,psfrag}
\usepackage{algorithm}
\usepackage[noend]{algorithmic}
\usepackage{mathtools,enumitem}
\usepackage{multirow}
\usepackage{ragged2e}
\usepackage{xr-hyper}
\usepackage{caption}
\usepackage{array}

\usepackage[utf8]{inputenc} 
\usepackage[T1]{fontenc}    
\usepackage{booktabs}       
\usepackage{nicefrac}         
\usepackage{microtype}      

\usepackage[dvipsnames]{xcolor} 
\definecolor{hunyuanblue}{HTML}{1E4A8F}
\usepackage{pifont} 
\usepackage{etoc} 
\usepackage{tikz} 
\usepackage{pgfplots}  
\pgfplotsset{compat=1.16}  

\ifdefined\TimesFont 
\usepackage{times} 
\fi

\ifdefined\ParSkip 
\usepackage{parskip} 
\fi

\newtheorem*{assumption*}{\assumptionnumber}
\providecommand{\assumptionnumber}{}


\makeatletter
\newcommand*\rel@kern[1]{\kern#1\dimexpr\macc@kerna}
\newcommand*\widebar[1]{%
  \begingroup
  \def\mathaccent##1##2{%
    \rel@kern{0.8}%
    \overline{\rel@kern{-0.8}\macc@nucleus\rel@kern{0.2}}%
    \rel@kern{-0.2}%
  }%
  \macc@depth\@ne
  \let\math@bgroup\@empty \let\math@egroup\macc@set@skewchar
  \mathsurround\z@ \frozen@everymath{\mathgroup\macc@group\relax}%
  \macc@set@skewchar\relax
  \let\mathaccentV\macc@nested@a
  \macc@nested@a\relax111{#1}%
  \endgroup
}
\makeatother

\def\E{\mathbb{E}}



%% file: _define.tex
\def\vc{{\bm{c}}}
\def\vv{{\bm{v}}}
\def\vx{{\bm{x}}}
\def\vz{{\bm{z}}}
\def\rmI{{\mathbf{I}}}
\def\gL{{\mathcal{L}}}
\def\gN{{\mathcal{N}}}
\def\gQ{{\mathcal{Q}}}
\def\gS{{\mathcal{S}}}
\def\gT{{\mathcal{T}}}

\newcommand{\ours}{FlowBP\xspace}
\newcommand{\oursleap}{FlowBP-Lagrange\xspace}
\newcommand{\ourscompose}{FlowBP-Sparse\xspace}
\newcommand{\ourscomposeleap}{FlowBP-Bridge\xspace}

\newcommand{\A}{\mathcal{A}}
\newcommand{\Apre}{\mathcal{A}_{\mathrm{pre}}}
\newcommand{\Apost}{\mathcal{A}_{\mathrm{post}}}

\providecommand{\vepsilon}{\bm{\epsilon}}
\providecommand{\sg}{\operatorname{sg}}

\newcommand{\xroll}{\vx_0}
\newcommand{\xhat}[1]{\hat{\vx}_0^{(#1)}}
\newcommand{\xsurrj}{\hat{\vx}_j}
\newcommand{\xrollj}{\vx_j}
\newcommand{\wfull}[1]{h_{#1}}
\newcommand{\M}{\mathcal{M}}
\newcommand{\Qleap}{\mathcal{Q}_{\mathrm{leap}}}
\newcommand{\Fnest}{\mathcal{F}_{\mathrm{nest}}}
\newcommand{\TargetFidelityAt}[1]{1-\frac{\E\!\left[\left\|\xhat{#1}-\xroll\right\|_2\right]}{\E\!\left[\left\|\xroll\right\|_2\right]}}
\newcommand{\RewardPathFidelity}{\mathcal{C}=\frac{\sum_{i\in\A} w_i\left(\TargetFidelityAt{i}\right)}{\sum_{i=1}^{N}\wfull{i}}}
\newcommand{\NestedDepth}{d_{\mathrm{nest}}(\M)}
\newcommand{\LeapQuality}{\Qleap=1-\frac{\E\!\left[\left\|\xsurrj-\xrollj\right\|_2\right]}{\E\!\left[\left\|\xrollj\right\|_2\right]}}
\newcommand{\NestedCouplingFidelity}{\Fnest=\frac{\NestedDepth}{N-1}\cdot\Qleap}

%% file: sec/00_abs.tex
Aligning text-to-image flow matching models with human preferences via direct reward backpropagation is sample-efficient but hampered by two well-known pathologies: activations cannot be stored across the full sampling trajectory at modern model scale, and chained Jacobian products across steps inflate the reward gradient as it travels back to early indices. Connector-based methods, such as LeapAlign, address these issues by replacing the full backward trajectory with a short pinned path, highlighting a useful decoupling between sampling and optimization. However, the quality of the resulting gradient depends on how accurately this short path approximates the full rollout, especially over long intervals. We propose \ours, a unified surrogate-trajectory framework that treats the backward trajectory itself as the design object. \ours keeps a no-gradient cached rollout for sampling, then builds a lightweight backward surrogate from cached and selectively re-forwarded velocities. This view separates four choices: the reward-model input, active set, integration weights, and bridge coupling, and recovers prior direct-gradient methods as particular settings. Within this framework, we instantiate three variants: \ourscompose uses sparse Euler reconstruction, \ourscomposeleap adds controlled bridge coupling, and \oursleap raises the order of leap quadrature. All three bound memory by the active-set size and limit gradient chaining to at most one Jacobian factor. Across \texttt{SD3.5-M}, \texttt{FLUX.1-dev}, and \texttt{FLUX.2-Klein-base} on preference, quality, and compositional metrics, the three variants improve over direct-gradient baselines on most metrics.

%% file: sec/01_intro.tex
\section{Introduction}
\label{sec:intro}

Pretrained text-to-image flow matching models~\citep{liu2023flow,lipman2023flow,esser2024scaling,flux2024,flux2klein} produce high-quality images, but are not aligned with human preferences. Two post-training paradigms have emerged to close this gap. \emph{Policy-gradient methods}~\citep{black2024training,liu2025flow,xue2025dancegrpo} convert the sampler into a stochastic policy and optimize a learned reward through likelihood-style estimators. \emph{Direct reward backpropagation}~\citep{xu2023imagereward,clark2024directly,wu2024drtune,liang2026leapalign} takes a different route: it exploits the differentiability of the flow sampler and propagates the reward gradient end-to-end through the same trajectory used at inference. The latter is sample-efficient and avoids the variance of stochastic policy gradients, but it comes with two well-known pathologies. Storing activations across the full sampling trajectory is infeasible at modern model scale, and the chained Jacobian product across steps inflates the reward gradient as it travels back~\citep{clark2024directly}.

At their root, the two pathologies leave any direct-gradient method facing the same four decisions: what input the reward model receives, which sampling steps carry the gradient, how their contributions are weighted, and whether a nested gradient path is retained and how it is scaled. Prior methods instantiate this surrogate in sharply different ways, but the underlying design space has not been explored. ReFL~\citep{xu2023imagereward} keeps a single late-stage step differentiable and evaluates the reward on a one-step Tweedie estimate. DRaFT-LV~\citep{clark2024directly} pins the differentiable step to the last index and reduces variance by averaging the reward gradient over multiple noise draws. DRTune~\citep{wu2024drtune} differentiates a sparse set of timesteps but stops the gradient at every intermediate input, deliberately removing the nested gradient path to keep the gradient bounded. LeapAlign~\citep{liang2026leapalign} replaces the long rollout with two single-velocity hops pinned by straight-through latent connectors, retaining a nested gradient path without storing the full trajectory. This connector-based construction highlights a useful decoupling between sampling and optimization; at the same time, its gradient quality depends on how accurately the short pinned path approximates the full rollout, especially over long intervals. Each method is effective in its own regime, yet their surrogate choices were designed in isolation, leaving the underlying trade-offs and unexplored configurations unclear.

We propose \ours{}, a unified surrogate-trajectory framework that treats the backward trajectory itself as the design object. \ours{} keeps a no-gradient cached rollout for sampling, then builds a lightweight backward surrogate from cached and selectively re-forwarded velocities. Differentiating this surrogate yields a single reward gradient governed by four design axes: the \emph{reward-model input}, the \emph{active set}, the \emph{integration weights}, and the \emph{bridge coupling}. The construction bounds the original pathologies directly: memory scales with the active-set size rather than the rollout length, and gradient inflation is limited to a single Jacobian factor instead of a multi-step product. It also recovers prior methods as particular settings of the same design space: ReFL, DRaFT-LV, and DRTune use short or sparse detached surrogates whose rewards are evaluated on posterior-mean estimates, while LeapAlign uses a connector-based pinned path that keeps the reward on the sampled image and retains one nested coupling path. This map exposes several unexplored surrogate designs, motivating the three instantiations below.

We instantiate three new methods in this design space (\cref{sec:method:instances}). \ourscompose{} uses sparse Euler reconstruction: it replays cached Euler updates over the rollout, exposes only a moderate active set to autograd, drops the bridge entirely, and yields a fully decoupled gradient with no cross-step Jacobian. \ourscomposeleap{} adds controlled bridge coupling by distributing a multi-step Euler active set across two segments joined by a bridge with tunable coupling strength. \oursleap{} raises the order of leap quadrature, replacing each one-velocity hop with a high-order Lagrange rule while keeping the two-segment leap structure. Evaluated on three flow-matching backbones across preference, quality, and compositional metrics, the three variants improve over the strongest direct-gradient baselines on most metrics, while their per-axis differences isolate each design choice's contribution.

Our contributions are as follows:
\begin{itemize}[leftmargin=*,itemsep=1pt,topsep=2pt]
    \item \ours{}, a unified surrogate-trajectory framework that treats the backward surrogate as the design object, subsumes prior direct-gradient methods along four axes, and bounds both memory and gradient by design.
    \item Three instantiations (\ourscompose{}, \ourscomposeleap{}, and \oursleap{}) that respectively explore sparse Euler reconstruction, controlled bridge coupling, and higher-order leap quadrature.
    \item Consistent gains across three flow-matching backbones on preference and quality metrics, together with improved compositional generation, achieving the strongest overall results among direct-gradient methods.
\end{itemize}

%% file: sec/06_related.tex
\section{Related Work}
\label{sec:related}

\noindent\textbf{Preference alignment for diffusion and flow models.}
Recent work explores diverse post-training strategies for aligning visual generative models with human preferences. Policy-gradient and GRPO-style methods adapt RL objectives to diffusion and flow-model sampling trajectories~\citep{black2024training,fan2023reinforcement,liu2025flow,xue2025dancegrpo,li2025mixgrpo}, while preference-optimization methods learn from paired, step-wise, dense, or self-generated preference signals~\citep{Wallace_2024_CVPR,yang2024using,Liang_2025_CVPR,yuan2024selfplay,yang2024adensereward,zhang2026diffusion}. Related likelihood- or score-based objectives further align diffusion models through score matching, discriminative optimization, or stochastic optimal control~\citep{zhu2025dspo,zheng2025direct,domingo-enrich2025adjoint}. Other flow or diffusion alignment methods connect online reward optimization with the model's pretraining or forward noising process~\citep{zheng2026diffusionnft,xue2025advantage}. These methods are broadly applicable to non-differentiable rewards, but they typically optimize stochastic trajectory objectives, preference likelihoods, control objectives, or likelihood-style surrogates rather than directly differentiating the reward through the sampler. Our work studies this complementary direct-gradient regime.

\noindent\textbf{Direct reward backpropagation.}
Direct-gradient methods exploit the differentiability of diffusion or flow samplers to optimize reward scores without likelihood estimation~\citep{prabhudesai2023aligning,xu2023imagereward,clark2024directly,wu2024drtune,zhang2025itercomp,shen2025directlyaligningdiffusiontrajectory,liang2026leapalign}. Existing approaches backpropagate rewards through short late-stage denoising paths, low-variance final-step objectives, detached intermediate inputs, composition-aware feedback, full-trajectory preference signals, or learned leap trajectories. These methods demonstrate the effectiveness of differentiable rewards, but each fixes a particular choice of reward input, active denoising steps, trajectory approximation, and nested-gradient scale. Our surrogate-trajectory view makes these choices explicit and uses them as design axes for the methods studied in this paper.

%% file: sec/02_background.tex
\section{Preliminaries}
\label{sec:preliminary}

\noindent \textbf{Diffusion and Flow models.}
Diffusion models~\citep{ho2020denoising,song2021scorebased} and flow matching~\citep{lipman2023flow,liu2023flow} are two formulations of the same continuous-time generative process. Under the velocity parameterization~\citep{salimans2022progressive}, both train a network $\vv_\theta(\vx_t, t, \vc)$ to predict the tangent of a trajectory $\vx_t = \alpha_t \vx_0 + \sigma_t \vepsilon$ that interpolates between clean data $\vx_0\sim p_{\text{data}}$ at $t=0$ and Gaussian noise $\vepsilon\sim\gN(\bm{0},\rmI)$ at $t=1$. Both sample by integrating the same probability-flow ODE $\mathrm{d}\vx_t/\mathrm{d}t = \vv_\theta(\vx_t, t, \vc)$; the two regimes differ only in the choice of noise schedule $(\alpha_t, \sigma_t)$. Our framework treats this schedule as a hyperparameter and applies to either regime, so for concreteness we work throughout in the rectified-flow setting~\citep{liu2023flow,esser2024scaling} ($\alpha_t = 1-t$, $\sigma_t = t$).

In this setting, the straight-line interpolation $\vx_t = (1-t)\,\vx_0 + t\,\vepsilon$ has target velocity $\vv = \mathrm{d}\vx_t/\mathrm{d}t = \vepsilon - \vx_0$, and the network is trained against this target by minimizing
\begin{equation}
\label{eq:fm_loss}
    \gL_{\text{FM}}(\theta)
    \;=\;
    \E_{t,\, \vx_0,\, \vepsilon}
    \!\left[\, \bigl\|\vv_\theta(\vx_t, t, \vc) - (\vepsilon - \vx_0)\bigr\|_2^2 \,\right].
\end{equation}

To generate samples, we discretize reverse time into an $N$-step rollout with schedule $\sigma_N = 1 > \sigma_{N-1} > \cdots > \sigma_0 = 0$, where $\sigma_i \in [0,1]$ is the time (equivalently, noise level) at step $i$; for example, a uniform discretization sets $\sigma_i = i/N$. Starting from initial noise $\vx_N \sim \gN(\bm{0}, \rmI)$, the Euler update
\begin{equation}
\label{eq:euler}
    \vx_{i-1} \;=\; \vx_i \;-\; (\sigma_i - \sigma_{i-1})\,\vv_i,
    \qquad \vv_i := \vv_\theta(\vx_i, \sigma_i, \vc),
    \qquad i = N, \ldots, 1,
\end{equation}
produces a trajectory $\{\vx_i\}_{i=N}^{0}$ that terminates at the clean sample $\vx_0$. Since the rollout is fully differentiable, reward gradients can be backpropagated through any subset of these $N$ steps to update the parameters $\theta$~\citep{xu2023imagereward,clark2024directly,wu2024drtune,liang2026leapalign}.

\noindent\textbf{Straight-through connectors.}
A common primitive used by trajectory-level reward methods such as LeapAlign~\citep{liang2026leapalign} is the \emph{straight-through connector}: at any rollout index $i$, given a differentiable surrogate estimate $\hat\vx_i$, the in-place pin
\begin{equation}
\label{eq:connector}
    \vx_i \;=\; \hat\vx_i + \sg(\vx_i - \hat\vx_i),
\end{equation}
keeps the forward value of $\vx_i$ equal to its cached value while rerouting
the gradient through $\hat\vx_i$, so that
$\partial\vx_i/\partial\theta = \partial\hat\vx_i/\partial\theta$. Following
LeapAlign, we continue to use the symbol $\vx_i$ for the post-connector latent.
We refer to $d_i=\|\vx_i-\hat\vx_i\|_2$ as the \emph{connector residual} at
index $i$: it does not affect the forward value but quantifies the discrepancy
between the forward and backward paths at the connection point.

%% file: fig/Flux1_overview_radar.tex
\begin{figure}[!h]
   \centering
   \begin{minipage}[b]{\linewidth}
       \begin{minipage}[b]{0.45\linewidth}
            \centering
            \includegraphics[width=1.00\textwidth]{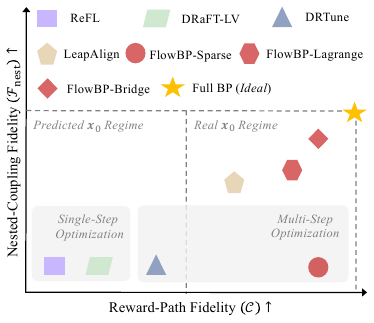}
       \end{minipage}\hfill
       \begin{minipage}[b]{0.53\linewidth}
            \centering
            \includegraphics[width=1.00\textwidth]{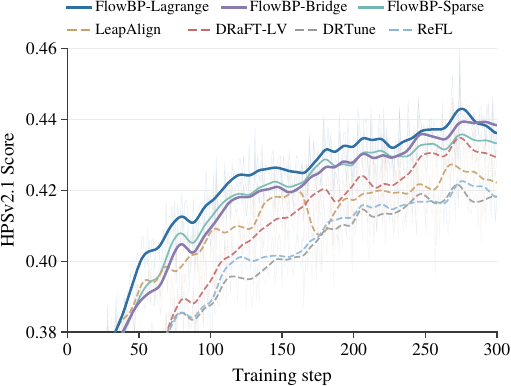}
       \end{minipage}
   \end{minipage}
    \caption{Overview of the \ours{} design space and empirical behavior.
    Left: methods are positioned by reward-path fidelity $\mathcal{C}$ and nested-coupling fidelity $\mathcal{F}_{\mathrm{nest}}$ (defined in \cref{app:surrogate_fidelity}), which respectively summarize reward-input fidelity and retained cross-step gradient coupling.
    Right: HPSv2.1 training reward on \texttt{FLUX.1-dev}; our three variants optimize faster and reach higher reward than the direct-gradient baselines.}
   \label{fig:flux1_overview_radar}
\end{figure}

%% file: sec/03_method.tex
\section{Method}
\label{sec:method}

\providecommand{\sg}{\operatorname{sg}}
\newcommand{\Dvt}[1]{\frac{\partial \vv_\theta(\vx_{#1})}{\partial \theta}}
\newcommand{\Dvx}[1]{\frac{\partial \vv_\theta(\vx_{#1})}{\partial \vx_{#1}}}

We aim to align a pretrained flow matching model with a scalar reward $r(\vx_0, \vc)$ by directly maximizing $\E[r(\vx_0, \vc)]$ via gradient ascent. The rollout in \cref{eq:euler} is fully differentiable, so $\partial r(\vx_0, \vc)/\partial \theta$ can in principle be obtained by reverse-mode autodiff through all $N$ Euler steps. In practice, naive backpropagation faces two well-known obstacles, and a third arises in the connector-based remedies prior work introduces to mitigate them (\cref{sec:method:setup}). We address all three with a unified \emph{surrogate-trajectory} framework (\cref{sec:method:framework}) that recovers ReFL~\citep{xu2023imagereward}, DRaFT-LV~\citep{clark2024directly}, DRTune~\citep{wu2024drtune}, and LeapAlign~\citep{liang2026leapalign} as special cases, motivating three new methods that improve on prior work along complementary axes: \ourscompose{}, \ourscomposeleap{}, and \oursleap{} (\cref{sec:method:instances}).

\subsection{Setup and Challenges}
\label{sec:method:setup}

Naively computing $\partial r(\vx_0, \vc)/\partial \theta$ by backpropagating through the Euler rollout in \cref{eq:euler} suffers from two well-documented challenges~\citep{clark2024directly,wu2024drtune,liang2026leapalign}, and a third challenge surfaces in the connector-based remedies that prior work introduces to mitigate them.

\noindent\textbf{Activation memory.} Each Euler step calls $\vv_\theta$, whose backward pass requires the corresponding activations to be retained. Storing all $N$ activations is prohibitive for modern backbones.

\noindent\textbf{Gradient explosion.} Backpropagation through the Euler chain in \cref{eq:euler} accumulates Jacobians $\bigl(\rmI - (\sigma_i - \sigma_{i-1})\,\Dvx{i}\bigr)$ across steps, which can amplify or attenuate gradients exponentially with the number of chained steps, making gradients for early-step updates unstable~\citep{clark2024directly}.

\noindent\textbf{Connector--induced mismatch.} LeapAlign~\citep{liang2026leapalign} constructs $\partial r(\vx_0, \vc)/\partial \theta$ by applying the straight-through connector~\eqref{eq:connector} at the rollout endpoint, but this splits the forward and backward paths there: $\nabla_\vx r$ is evaluated at $\vx_0$ yet composed with $\partial \hat\vx_0/\partial \theta$, where $\hat\vx_0$ is the connector surrogate endpoint. The mismatch scales with the connector residual $d_0=\|\vx_0-\hat\vx_0\|_2$, which grows when a single velocity spans a long interval; once $d_0$ becomes large, the surrogate gradient is no longer a reliable update direction and training destabilizes (\cref{fig:d0_reward_dynamics}).

\subsection{Unified Surrogate-Trajectory Framework}
\label{sec:method:framework}

LeapAlign shows that straight-through connectors can avoid full-trajectory backpropagation, but its long single-velocity leaps can leave large connector residuals and route reward gradients through a biased surrogate. We address the memory, gradient-chain, and connector-mismatch issues in one surrogate-trajectory construction, reducing the mismatch either \emph{numerically} with higher-order leap quadrature or \emph{structurally} through Euler reconstruction that matches the rollout endpoint. We run the Euler solver in \cref{eq:euler} without autograd and cache the resulting states and velocities $\{(\vx_i,\vv_i)\}_{i=0}^N$; rather than backpropagating through this full rollout, we construct a lightweight surrogate trajectory on top of the cache.

The surrogate shares its forward values with the cached rollout but carries a different backward graph. Only a small subset of velocity evaluations are re-forwarded with gradients, while all other states and velocities are treated as constants. The reward is evaluated on the original rollout sample, whereas gradients flow through a sparse surrogate graph. As a result, memory scales with the number of active velocities rather than the rollout length $N$, and gradient propagation no longer traverses the full Euler chain.

For an overview of the framework and its relationship to prior methods, \cref{fig:pipeline} visualizes the corresponding surrogate graphs.

\input{A-fig/Pipeline}

\noindent\textbf{Surrogate structure.}
The surrogate approximates the cached rollout using one or two sparse segments. We use $k$ for the noisy-side anchor of a connector-style segment and $j$ for an optional split index, with
\[
N \ge k > j \ge 0.
\]
When $j=0$, the surrogate has a single segment ending at $\vx_0$; connector-style methods may start from $\vx_k$, while Euler-reconstruction methods set $k=N$ and replay the cached rollout from $\vx_N$. When $j>0$, the surrogate has a pre-segment and a post-segment,
\[
\vx_k \;\rightarrow\; \vx_j \;\rightarrow\; \vx_0,
\]
with the reconstruction-based bridge using the special case $k=N$, i.e., $\vx_N\rightarrow\vx_j\rightarrow\vx_0$.

\noindent\textbf{Reward-model input.}
This axis specifies only where the scalar reward is queried. Detached short surrogates evaluate $r$ on a posterior-mean estimate $\hat\vx_0$, whereas endpoint-faithful variants make the reward read the cached rollout sample $\vx_0$. There are two ways to obtain this endpoint-faithful forward value without full-trajectory backpropagation: a straight-through endpoint connector, which keeps the forward value at $\vx_0$ but differentiates through $\hat\vx_0$ and is therefore sensitive to the endpoint residual $d_0=\|\vx_0-\hat\vx_0\|_2$; or Euler reconstruction, which replays the cached discrete updates while replacing only active velocities by re-forwarded copies, reproducing $\vx_0$ by construction and eliminating endpoint connector bias.

\noindent\textbf{Bridge coupling.}
This axis applies only when a split index $j$ is used, and controls whether gradients from the post-segment are allowed to affect the pre-segment through the split latent. We denote the nested-gradient scale by $\alpha\in[0,1]$. With no bridge ($j=0$), the surrogate contains only direct active-step terms. With a bridge, the nested-gradient scale $\alpha$ determines how much gradient crosses the split: $\alpha=0$ detaches the two sides, while a larger $\alpha$ supplies controlled cross-segment credit assignment through a single Jacobian. The bridge latent itself can be formed either by a connector, whose quality depends on the split residual $d_j=\|\vx_j-\hat\vx_j\|_2$, or by Euler reconstruction, which matches the cached $\vx_j$ in the forward pass and avoids a split-connector residual. Concrete realizations are given in \cref{sec:method:instances}.

\noindent\textbf{Active set.}
Only a subset of velocities participates in gradient computation. We denote this active set by
\[
\A = \Apre \cup \Apost,
\]
where $\Apre \subseteq (j,k]$ (with $k=N$ for full Euler reconstruction) and $\Apost \subseteq (0,j]$ correspond to the pre- and post-segments, respectively. We write $K=|\A|$ for the number of active velocities.

Each cached velocity is replaced with one of three values:
\begin{equation}
\label{eq:surrogate_v}
    \tilde\vv_i \;=\;
    \begin{cases}
        \vv_\theta\bigl(\alpha\,\vx_j + (1-\alpha)\,\sg(\vx_j),\, \sigma_j,\, \vc\bigr), & i = j, \\[2mm]
        \vv_\theta\bigl(\sg(\vx_i),\, \sigma_i,\, \vc\bigr), & i \in \A,\, i \ne j, \\[2mm]
        \sg(\vv_i), & i \notin \A.
    \end{cases}
\end{equation}

Active steps are re-forwarded so gradients flow through $\theta$, while their inputs remain detached to prevent gradient propagation into earlier rollout states. Inactive steps reuse the cached velocity directly. The bridge step uses the split latent $\vx_j$ and the coupling scale $\alpha$ to control whether post-segment gradients flow back into the pre-segment.

\noindent\textbf{Surrogate trajectory.}
The cached rollout gives us the reference states $\vx_k,\vx_j,\vx_0$. The surrogate graph can connect to these states in two ways. The first is a leap estimate: for an interval $s\rightarrow t$, \cref{eq:leap_estimate} predicts
 an approximate target and then pins it back to the cached state with the connector in \cref{eq:connector}:
\begin{equation}
\label{eq:leap_estimate}
    \hat\vx_t
    =
    \vx_s-\sum_{i\in [t+1,s]} w_i\,\tilde\vv_i,
    \qquad
    \vx_t
    =
    \hat\vx_t+\sg(\vx_t-\hat\vx_t).
\end{equation}
\Cref{eq:leap_estimate} is lightweight, but if $\hat\vx_t$ differs from the cached $\vx_t$, the connector residual becomes part of the surrogate bias. The second is a reconstruction estimate: it replays the discrete Euler interval itself,
\begin{equation}
\label{eq:euler_reconstruction}
    \vx_t
    =
    \vx_s-\sum_{i\in [t+1,s]} h_i\,\tilde\vv_i,
    \qquad h_i=\sigma_i-\sigma_{i-1}.
\end{equation}
Because inactive velocities are the cached velocities and active velocities are re-forwarded at the same cached states, the reconstruction in \cref{eq:euler_reconstruction} has the same forward value as the original rollout state. Thus, reconstruction keeps the sparse backward graph but does not need a connector residual to make the reward or bridge latent lie on the cached trajectory.

\noindent\textbf{Unified gradient.}
At the gradient level, both constructions expose the same two kinds of terms:
\begin{equation}
\label{eq:unified_grad}
    \frac{\partial\vx_0}{\partial\theta}
    \;=\;
    \underbrace{
    -\sum_{i\in\A} w_i\,\Dvt{i}
    }_{\text{direct terms}}
    \;+\;
    \underbrace{
    \alpha\,w_j\,\Dvx{j}
    \sum_{i\in\Apre} w_i\,\Dvt{i}
    }_{\text{nested term}}.
\end{equation}

The direct terms update each active step independently. The nested term is the only place where a Jacobian re-enters the gradient, and its depth is exactly one regardless of the rollout length $N$. Consequently, memory scales with $|\A|$ rather than $N$, while gradient amplification is bounded by a single Jacobian factor rather than a multi-step product.

\noindent\textbf{Design axes.}
The unified gradient reveals four independent design choices:

\begin{itemize}[leftmargin=1.2em,itemsep=1pt,topsep=2pt]
\item \textbf{Reward-model input}: whether the reward evaluates the rollout sample $\vx_0$ or an estimate $\hat\vx_0$.
\item \textbf{Active set}: which velocities are re-forwarded with gradients.
\item \textbf{Integration weights}: how active velocities are combined through $\{w_i\}$.
\item \textbf{Bridge coupling}: whether a split index $j$ is introduced, how the split latent is formed, and how strongly gradients couple across it through $\alpha$.
\end{itemize}

As shown in \cref{tab:unified_design}, existing methods correspond to particular settings of these four axes, while our proposed variants explore previously unexplored combinations.

\subsection{Method Instantiations}
\label{sec:method:instances}

\cref{tab:unified_design} maps each method onto the four axes, and prior methods cluster into two regimes. ReFL, DRaFT-LV, and DRTune never reconstruct the rollout endpoint, so the reward model reads the posterior-mean estimate $\hat\vx_0$, which lies off the natural-image manifold; with no bridge, the nested term in \cref{eq:unified_grad} vanishes, and the gradient collapses to a short direct sum with no nested gradient path. LeapAlign is the only prior method that pins both a bridge and an endpoint with connectors, but its single-velocity quadrature can incur large connector residuals over long leap intervals.

\input{tab/method_compare}

\noindent\textbf{ReFL~\citep{xu2023imagereward}.}
A single-step active set $\A = \{\tau\}$ with weight $w_{\tau} = \sigma_{\tau}$ at a randomly sampled index $\tau$ corresponds to a one-step Tweedie jump $\hat\vx_0 = \vx_{\tau} - \sigma_{\tau} \vv_{\tau}$ from $\vx_{\tau}$ along the predicted velocity. In our framework, no endpoint or bridge connector is introduced: the reward model sees the posterior-mean estimate $\hat\vx_0 = \E[\vx_0\mid\vx_{\tau}]$, and the gradient reduces to a single direct factor $-\sigma_\tau\,\Dvt{\tau}$ with no nested gradient path.

\noindent\textbf{DRaFT-LV~\citep{clark2024directly}.}
DRaFT-LV uses the same single-step active set as ReFL but pins it to the last index ($\A = \{1\}$, $w_1 = \sigma_1$), with variance reduced by averaging $\nabla_x r$ over multiple noise draws at $\vx_1$. Again, no endpoint or bridge connector is introduced; the reward model sees $\hat\vx_0 = \E[\vx_0\mid\vx_1]$, and the nested term vanishes, leaving a single direct factor.

\noindent\textbf{DRTune~\citep{wu2024drtune}.}
DRTune broadens the active set to a sparse Euler quadrature over a sampled subset $t_{\mathrm{train}}$ of training timesteps plus a final Tweedie step at $t_{\min}$, with Euler weights $h_i$ on $t_{\mathrm{train}}$ and weight $\sigma_{t_{\min}}$ on the Tweedie step. No endpoint or bridge connector is introduced, and the gradient is a $K$-summand direct sum; the reward model sees $\hat\vx_0 = \E[\vx_0\mid\vx_{t_{\min}}]$. DRTune therefore broadens the active set relative to ReFL and DRaFT-LV but retains their connector-free structure.

\noindent\textbf{LeapAlign~\citep{liang2026leapalign}.}
LeapAlign is the only prior method to engage both connectors. Each segment uses a single velocity to span the full leap interval: $w_k = \sigma_k - \sigma_j$ on the pre-segment with $\Apre = \{k\}$, and $w_j = \sigma_j$ on the post-segment with $\Apost = \{j\}$. It uses the nested-gradient scale $\alpha \in (0,1]$ to weight the nested term. The reward sees the actual sample $\vx_0$, but $K = 2$ restricts the trajectory information the surrogate can carry. Consequently, the single-velocity bridge can produce large connector residuals $d_j$ and $d_0$ over long leap intervals; when these residuals spike, the reward gradient is routed through an inaccurate bridge or endpoint estimate, making training less stable. We empirically verify the endpoint-residual failure mode in \cref{fig:d0_reward_dynamics}.

Our three methods all have the reward model evaluate the actual rollout sample $\vx_0$ rather than a posterior mean estimate $\hat\vx_0$, but they achieve this goal in two complementary ways. We first develop reconstruction-based variants: \ourscompose{} reconstructs the endpoint by Euler composition while dropping the bridge, and \ourscomposeleap{} additionally reconstructs the bridge with a tunable nested-gradient scale. These methods remove connector residuals structurally. We then return to the connector-based path and ask how much potential remains if we improve LeapAlign's connector rather than replace it. This motivates \oursleap{}, which keeps the two-segment connector topology but reduces $d_j$ and $d_0$ numerically by increasing both the active support and the quadrature order within each leap. We develop each method in this order below.

\noindent\textbf{\ourscompose{}.}
\ourscompose{} uses the original Euler quadrature over the full rollout, while exposing only randomly sampled steps to autograd. In this case $j=0$, $\Apre=\A$, and $\Apost=\emptyset$, so the composed endpoint is
\begin{equation}
\label{eq:compose_surrogate}
    \vx_0 \;=\; \vx_N \;-\; \sum_{i=1}^{N} h_i\, \tilde\vv_i,
    \qquad h_i = \sigma_i - \sigma_{i-1},
\end{equation}
The unified gradient in \cref{eq:unified_grad} collapses to a clean sum,
\begin{equation}
\label{eq:compose_grad}
    \frac{\partial \vx_0}{\partial \theta}
    \;=\;
    -\sum_{i \in \A} h_i\, \Dvt{i}\,,
\end{equation}
with no nested term. Two properties follow: (i)~memory is $O(K)$ regardless of $N$; (ii)~the gradient is bounded by $K$ summands of bounded individual norm, eliminating gradient explosion by construction. \ourscompose{} thus shares the nested-free structure of ReFL, DRaFT-LV, and DRTune, but unlike them, it reconstructs the actual rollout endpoint by full Euler composition. Hence, the reward model reads the in-distribution sample $\vx_0$ without bias.

\noindent\textbf{\ourscomposeleap{}.} \ourscomposeleap{} combines the full Euler-quadrature trajectory of \ourscompose{} with a structurally exact bridge. A split index $j \in \{1, \ldots, N-1\}$ partitions the rollout into a pre-segment $(N\rightarrow j)$ and a post-segment $(j\rightarrow0)$, and the $K$ active slots are allocated approximately in proportion to segment length, with at least one active velocity on each side. Both segments use Euler weights $w_i = h_i$. The pre-segment composition reconstructs $\vx_j$ in the forward pass, and the velocity at the split index is re-forwarded on the discounted input from the $i = j$ branch of \cref{eq:surrogate_v} with $\alpha \in (0,1]$. The unified gradient in \cref{eq:unified_grad} retains both the direct sum (over both segments) and the nested term (through the bridge Jacobian $\Dvx{j}$):
\begin{equation}
\label{eq:composesplit_grad}
    \frac{\partial \vx_0}{\partial \theta}
    \;=\;
    -\sum_{i \in \A} h_i\, \Dvt{i}
    \;+\;
    \alpha\, h_j\, \Dvx{j} \sum_{i \in \Apre} h_i\, \Dvt{i}.
\end{equation}
\ourscomposeleap{} thus interpolates between \ourscompose{} (no bridge, fully decoupled gradient) and a one-Jacobian bridged update, trading some of \ourscompose{}'s decoupling guarantee for explicit nested coupling that propagates reward signals across the trajectory split. Unlike LeapAlign, this bridge is produced by Euler composition rather than by a biased single-velocity connector. Full derivations are in \cref{app:compose_split}.

\noindent\textbf{\oursleap{}.}
\oursleap{} extends LeapAlign by replacing each single-velocity leap with a Lagrange quadrature rule. The two leaps span $(j, k]$ and $(0, j]$ with support sets $\mathcal{S}_{\mathrm{pre}}$ and $\mathcal{S}_{\mathrm{post}}$, whose anchors are $k\in\mathcal{S}_{\mathrm{pre}}$ and $j\in\mathcal{S}_{\mathrm{post}}$. The gradient-active sets satisfy $\Apre\subseteq\mathcal{S}_{\mathrm{pre}}$ and $\Apost\subseteq\mathcal{S}_{\mathrm{post}}$, and the active count $K=|\Apre\cup\Apost|$ stays within a fixed active-support budget $M$ (i.e. $K\le M$). Integrating the segment-wise Lagrange basis polynomials gives weights $w_i^{\mathrm{L}}$ for each support; for the bridge anchor, $w_j^{\mathrm{L}}$ denotes the post-segment weight. Inactive supports, when present, are used only in the forward quadrature with detached cached velocities. The one-support-per-segment case recovers LeapAlign~\citep{liang2026leapalign}. The resulting gradient is
\begin{equation}
\label{eq:lagrange_grad}
    \frac{\partial \vx_0}{\partial \theta}
    =
    -\sum_{i \in \Apre}
        \rho_i\, w_i^{\mathrm{L}}\, \Dvt{i}
    -\sum_{i \in \Apost}
        \rho_i\, w_i^{\mathrm{L}}\, \Dvt{i} \\
    \quad
    +\alpha\, w_j^{\mathrm{L}}\,
        \Dvx{j}
        \sum_{i \in \Apre}
        \rho_i\, w_i^{\mathrm{L}}\, \Dvt{i}.
\end{equation}
Here $\rho_i=1$ for the start anchor of each segment, while non-anchor active supports are attenuated by a fixed gradient-support scale $g_s\in[0,1]$ (i.e.\ $\rho_i=g_s$) to keep the high-order correction from inflating the gradient norm. The two sums give the direct high-order support updates, and the last term is the same single-Jacobian bridge correction as in LeapAlign, now driven by the higher-order pre-segment quadrature. Full definitions of the weights and gradient-support policy are given in \cref{app:high_order_leapalign}.

\noindent\textbf{Design spectrum.} Our three methods separate into two ways of avoiding poor reward inputs. \oursleap{} stays close to LeapAlign and reduces connector bias by raising the quadrature accuracy under a compact active-support budget, improving the per-segment integration accuracy without inflating $K$ beyond $M$. \ourscompose{} and \ourscomposeleap{} instead take a structural route to mitigating connector bias: \ourscompose{} increases active-step coverage while using exact Euler composition and no bridge, trading the nested gradient path for unconditional stability; \ourscomposeleap{} combines a high-budget Euler active set with a structurally exact bridge and a tunable nested-gradient scale, retaining both the dense gradient signal of \ourscompose{} and a bounded one-step coupling path. We empirically map all three directions in \cref{sec:experiments}.

%% file: A-fig/Pipeline.tex
\begin{figure}[t]
\centering
\includegraphics[width=\textwidth]{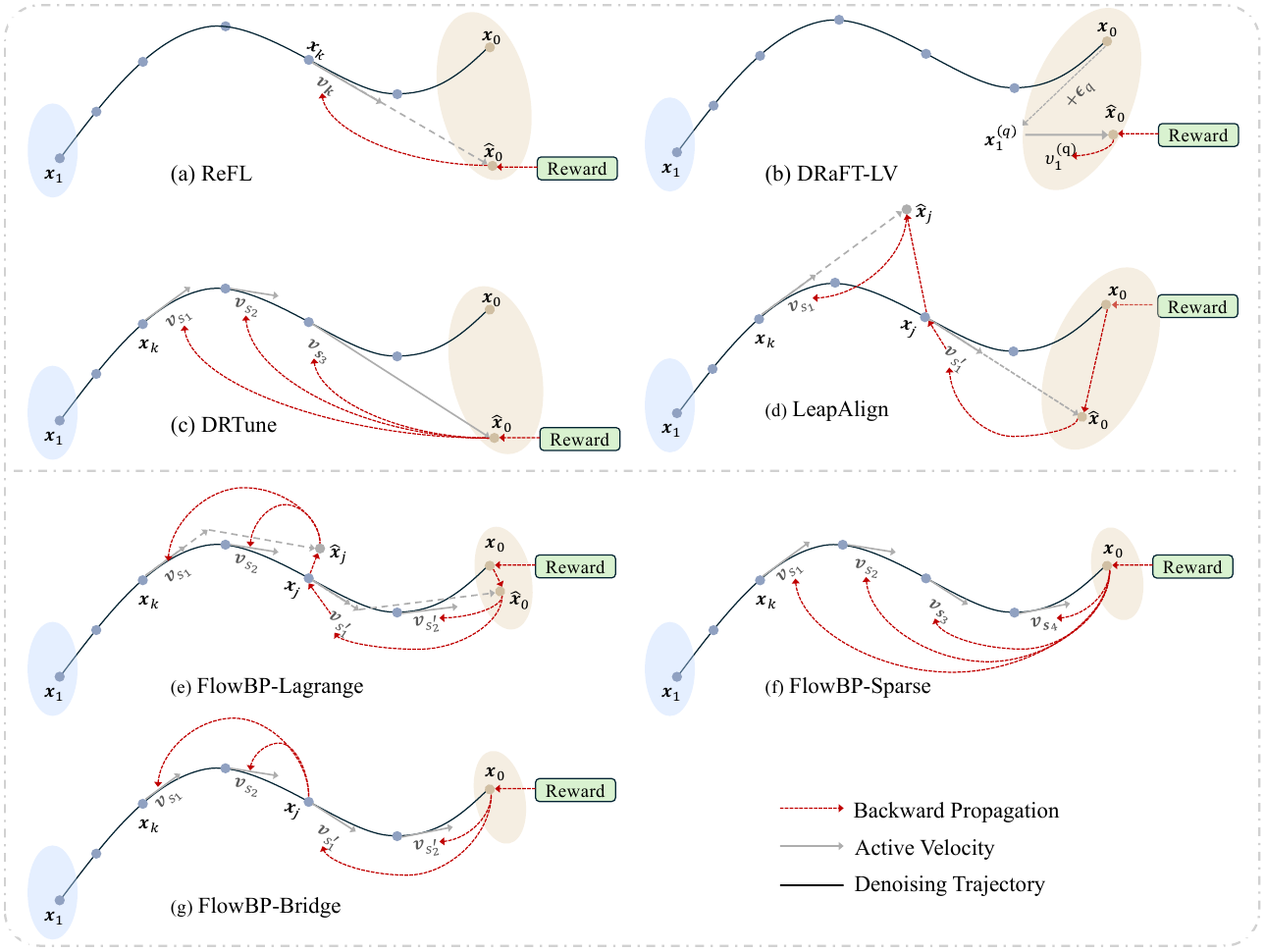}
\caption{Surrogate backward graphs under the unified framework. Each panel uses the same cached denoising trajectory but differs in reward-model input, active velocities, integration weights, and nested-gradient path. Highlighted velocities are re-forwarded with gradients, while inactive trajectory states and velocities are detached. ReFL, DRaFT-LV, and DRTune optimize posterior-mean endpoints with no nested path; LeapAlign keeps a two-step bridged path; \oursleap{} raises the bridge quadrature order, \ourscompose{} uses a sparse Euler active set without a bridge, and \ourscomposeleap{} combines sparse Euler coverage with a tunable nested-gradient scale.}
\label{fig:pipeline}
\end{figure}

%% file: tab/method_compare.tex
\begin{table}[t]
\centering
\caption{Instantiations of the unified surrogate-trajectory framework. \textbf{RM input} denotes the reward-model input, \textbf{Weights} lists the surrogate integration coefficients $w_i$, and \textbf{Bridge} reports the bridge setting: $j=0$ means no bridge, while $\alpha\in(0,1]$ sets the nested-gradient scale when a bridge is used. Here $K$ is the cardinality of the active set ($|\A|$), $h_i=\sigma_i-\sigma_{i-1}$, and $w_i^{\mathrm{L}}$ denotes the Lagrange quadrature weights.}
\label{tab:unified_design}
\setlength{\tabcolsep}{2.5pt}
\begin{tabular}{l|cccc}
\toprule
\textbf{Method} & \textbf{RM input} & \textbf{Weights} $w_i$ & \textbf{Active set} $\A$ & \textbf{Bridge} \\
\midrule
ReFL~\citep{xu2023imagereward} & $\E[\vx_0|\vx_\tau]$ & $w_\tau = \sigma_\tau$ & $\{\tau\}$ & $j{=}0$ \\
DRaFT-LV~\citep{clark2024directly} & $\E[\vx_0|\vx_1]$ & $w_{1} = \sigma_{1}$ & $\{1\}$ & $j{=}0$ \\
DRTune~\citep{wu2024drtune} & $\E[\vx_0|\vx_{t_{\min}}]$ & Euler $h_i$ and leap $\sigma_{t_{\min}}$ & $t_{\mathrm{train}} \cup \{t_{\min}\}$ & $j{=}0$ \\
\midrule
LeapAlign~\citep{liang2026leapalign} & $\vx_0$ & $w_{k} = \sigma_k{-}\sigma_j$, $w_j = \sigma_j$ & $\{k, j\}$, $k>j$ & $\alpha\in(0,1]$ \\
\ourscompose{} (ours) & $\vx_0$ & $h_i$ (Euler) & $K$ indices & $j{=}0$ \\
\ourscomposeleap{} (ours) & $\vx_0$ & $h_i$ (Euler) & $K$ indices, split at $j$ & $\alpha\in(0,1]$ \\
\oursleap{} (ours) & $\vx_0$ & $w_i^{\mathrm{L}}$ (Lagrange) & $K\le M$ & $\alpha\in(0,1]$ \\
\bottomrule
\end{tabular}
\end{table}

%% file: sec/04_experiment.tex
\section{Experiments}
\label{sec:experiments}

\subsection{Setup}
We evaluate our methods on three text-to-image flow-matching backbones spanning
different model families and capacities: \texttt{SD3.5-M}~\citep{esser2024scaling},
\texttt{FLUX.1-dev} ~\citep{flux2024}, and \texttt{FLUX.2-Klein-base} (9B)~\citep{flux2klein}. For each backbone, we report the base model and four reward-gradient post-training baselines:
ReFL~\citep{xu2023imagereward}, DRaFT-LV~\citep{clark2024directly},
DRTune~\citep{wu2024drtune}, and LeapAlign~\citep{liang2026leapalign}. We
implement these baselines following their official papers.

Following MixGRPO~\citep{li2025mixgrpo}, we evaluate all methods on the HPDv2
test split, which contains 400 held-out prompts. We use
HPSv2.1~\citep{wu2023human}, PickScore~\citep{kirstain2023pickapic}, and
ImageReward~\citep{xu2023imagereward} to measure human preference alignment, and
UnifiedReward-Alignment (UR-Align) and UnifiedReward-IQ (UR-IQ)~\citep{unifiedreward} to assess image-text alignment and
overall image quality. For each metric, we report the average score over the 400
prompts and the improvement over the corresponding base model. Since post-training
uses HPSv2.1 as the differentiable reward, we treat HPSv2.1 as the in-domain
metric and the remaining reward models as out-of-domain evaluators.

We further evaluate compositional generation on \texttt{FLUX.1-dev} using
GenEval~\citep{ghosh2023geneval}, following its official protocol, and report
task accuracies and the overall score computed by the official evaluator.

\subsection{Implementation Details}
For our variants, we follow the LeapAlign training protocol and optimize against
HPSv2.1~\citep{wu2023human} as the differentiable reward model, using a hinge
threshold of $\lambda=0.55$. Training prompts are sampled from HPDv2. Online
rollouts are generated at $512\times512$ resolution with 25 sampling steps. We
use classifier-free guidance~\citep{ho2022classifierfree} scale 3.5 for
\texttt{SD3.5-M} and 4.0 for both FLUX backbones. We train \texttt{SD3.5-M} for
250 iterations and both FLUX backbones for 300 iterations. We optimize the DiT
parameters with AdamW~\citep{loshchilov2019decoupled}, using learning rate
$1\times10^{-5}$, weight decay $10^{-4}$, EMA decay 0.995, and batch size 64. As
in LeapAlign, we use mixed-precision training with bf16 activations and maintain
master weights in fp32. At evaluation time, each checkpoint is sampled with 50
steps using the same resolution and guidance scale as training. Additional implementation details are provided in \cref{appsec:experimental_details}.

\subsection{Main Results}
\input{tab/main_experiment}
\input{fig/Flux1_image_comparsion}
\input{tab/geneval}
\FloatBarrier
\input{fig/Flux1_highorder_ablation}
\input{fig/ComposeLeap_ablation}
\cref{tab:main_experiment} reports the main comparison across the three
flow-matching backbones. Across all of them, one of our variants takes the best score
on nearly every preference and quality metric, with the lead rotating among the
three rather than concentrating on any single variant. On \texttt{SD3.5-M},
\oursleap is strongest on HPSv2.1, PickScore, ImageReward, and UR-Align, while
\ourscompose gives the best UR-IQ. On \texttt{FLUX.1-dev}, \oursleap leads on
HPSv2.1 and UR-Align, whereas \ourscomposeleap leads on PickScore, ImageReward,
and UR-IQ. On \texttt{FLUX.2-Klein-base}, the reconstruction-based variants dominate: \ourscomposeleap takes the best HPSv2.1 and \ourscompose the best PickScore,
ImageReward, and UR-IQ. The one metric where a prior method retains the lead is UR-Align on
\texttt{FLUX.2-Klein-base}, led by LeapAlign. Every variant improves over its
base model on all five metrics, and no single variant wins everywhere; the
lead shifts with backbone and reward dimension, indicating that the three variants
capture genuinely different aspects of the reward landscape rather than
redundant gains.

\cref{tab:geneval_flux1} evaluates compositional generation on GenEval. On
\texttt{FLUX.1-dev}, \oursleap improves the overall score from 63.25 to 69.88,
achieving the best overall performance and the best scores on three categories.
\ourscomposeleap is second overall and performs best on spatial position. These results indicate that the preference improvements do not degrade compositional
prompt following.

\cref{fig:flux1_overview_radar} summarizes the design space view together with
the \texttt{FLUX.1-dev} training dynamics: reward improves stably throughout optimization. \cref{fig:flux1_geneval_qualitative} shows representative GenEval examples consistent with the quantitative trends in \cref{tab:geneval_flux1}. We also note that the strongest variant differs across backbones and metrics, suggesting that the surrogate-trajectory design interacts with model scale and architecture. Per-backbone evaluation dynamics over the course of training are reported in \cref{fig:app_flux_eval_curves}, and additional qualitative results across all three backbones are provided in \cref{appsec:qualitative_results}.

\subsection{Ablation Studies}

We use \texttt{FLUX.1-dev} as the ablation backbone because it provides stable
training dynamics and a representative model scale. We ablate the four design
axes from \cref{sec:method:framework} through the concrete knobs used by our
variants: the reward-model input via endpoint reconstruction, the active set
by varying the budget $K$, the integration weights via the different quadrature rules, and the bridge coupling via the nested-gradient scale $\alpha$.
\input{fig/Loss_d0_comparsion}
\input{A-tab/endpoint_connector_ablation}

\noindent\textbf{Reward-model input.}
To isolate the reward-model input, we replace the posterior-mean endpoint used by
ReFL and DRTune with the same Euler-composed endpoint treatment as \ourscompose{}: the cached rollout is replayed with the selected active velocities re-forwarded, so the forward endpoint equals the sampled image $\vx_0$ by construction. This is not a straight-through endpoint connector and therefore does not introduce an endpoint pinning residual; it gives an endpoint-faithful surrogate with respect to the cached Euler rollout while keeping the active set, weights, and optimization fixed. This change improves every metric on \texttt{FLUX.1-dev} (\cref{tab:endpoint_reconstruction}), showing that placing the reward on the actual rollout sample is important even before changing the active set or quadrature rule.

\noindent\textbf{\oursleap connector design.}
\cref{fig:flux1_highorder_ablation} studies the two main choices in \oursleap: the Lagrange quadrature
order and the gradient-support scale. All high-order variants use the same support points within each leap; the uniform connector simply assigns equal coefficients to these supports, whereas the Lagrange connector computes the coefficients by integrating the Lagrange basis. The Lagrange connector attains much lower endpoint-prediction error than both the single-anchor Euler connector and the equal-weight uniform connector (right), with
the gap widening over long leaps where uniform averaging still fails to track trajectory curvature. This directly explains the higher, more stable reward (left). The support scale must be balanced (middle): too small reverts to single-velocity LeapAlign, while too large overweights the high-order corrections and destabilizes training. This motivates our default, Lagrange connector with a moderate support scale; the weights and attenuation are derived in \cref{app:high_order_leapalign}.

\noindent\textbf{\ourscomposeleap{} nested-gradient scale.}
\cref{fig:composeleap_ablation} evaluates the two knobs of \ourscomposeleap. The nested-gradient scale $\alpha$ interpolates from near full decoupling ($\alpha\!=\!0$, recovering \ourscompose) toward stronger nested coupling. The reward peaks at an intermediate value: moderate nested coupling propagates useful signal across the split, while strong coupling reintroduces the single-Jacobian amplification that the bridge is meant to bound. The active-step budget $K$ controls how many denoising velocities are re-forwarded in the composed segment. The ablation shows that \ourscomposeleap benefits from a compact but nontrivial budget, confirming that it is most useful as a controlled middle ground rather than at either extreme (gradient expression in \cref{app:compose_split}).

\noindent\textbf{\ourscompose{} active-step budget.}
In \cref{fig:compose_ablation}, we vary the active-step budget $K$, i.e., the number of
re-forwarded velocities used in the Euler composition. Increasing $K$ gives denser
trajectory coverage but also increases the size of the backward graph. The ablation
identifies a compact budget that preserves reward gains while keeping the surrogate
sparse, justifying our default \ourscompose{} setting
(\cref{app:composed_velocity}).

\noindent\textbf{Connector deviation analysis.}
\cref{fig:d0_reward_dynamics} explains the benefit of high-order connectors. The
single-velocity leap in LeapAlign incurs a large connector deviation $d_0$, whose
spikes coincide with reward collapse (shaded interval): once the surrogate
endpoint drifts from the actual sample, the reward gradient is evaluated at an unreliable point, and training destabilizes. \oursleap keeps $d_0$ smaller via its
more accurate connector, so the endpoint tracks the actual sample and training
avoids this failure mode, linking its gains to a concrete, measurable mechanism.

\noindent\textbf{Design takeaways.}
The ablations clarify how the four surrogate-design axes interact in practice.
First, the reward-model input should be endpoint-faithful whenever possible:
using the same endpoint reconstruction as \ourscompose{} already improves ReFL
and DRTune across all metrics (\cref{tab:endpoint_reconstruction}). Second, the
integration weights matter over long surrogate intervals. \oursleap{} reduces
connector mismatch with a high-order quadrature rule and avoids the large $d_0$
failure mode observed in LeapAlign (\cref{fig:d0_reward_dynamics}). Third, the
active set should provide enough trajectory coverage without making the backward
graph unnecessarily large; the $K$ ablations for \ourscompose{} and
\ourscomposeleap{} both favor compact but nontrivial active budgets. Finally,
bridge coupling is useful only when controlled: \ourscompose{} shows that
detached multi-step updates are stable, while \ourscomposeleap{} shows that a
single $\alpha$-scaled nested Jacobian can add useful cross-step credit assignment
without recreating the instability of full backpropagation. Overall, a robust
surrogate should place the reward on the sampled endpoint, use accurate
integration over long intervals, expose a compact active set, and add bounded
bridge coupling only when an additional cross-step signal is needed.

%% file: tab/main_experiment.tex
\providecommand{\best}[1]{\textbf{#1}}
\providecommand{\second}[1]{\underline{#1}}
\definecolor{deltaGray}{gray}{0.45}
\definecolor{colSky}{RGB}{135,206,235}
\definecolor{groupShade}{RGB}{238,238,238}
\providecommand{\dt}[1]{}
\renewcommand{\dt}[1]{\textcolor{deltaGray}{\scriptsize #1}}

\begin{table}[!htbp]
\centering
\caption{
Main results on three text-to-image flow-matching backbones, evaluated with HPSv2.1, PickScore, ImageReward, UR-Align, and UR-IQ. $\Delta$ denotes the improvement over the corresponding base model. Best and second-best results are shown in \textbf{bold} and \underline{underlined}, respectively. Blue rows indicate our methods.
}
\label{tab:main_experiment}
\vspace{3pt}
\setlength{\tabcolsep}{1.7pt}
\renewcommand{\arraystretch}{1.13}
\resizebox{0.95\textwidth}{!}{
  \begin{tabular}{@{}l cc cc cc cc cc@{}}
  \toprule
  \multirow{3}{*}{\textbf{Method}} &
  \multicolumn{2}{c}{\textbf{In-Domain}} &
  \multicolumn{8}{c}{\textbf{Out-of-Domain}} \\
  \cmidrule(lr){2-3} \cmidrule(lr){4-11}
  &
  \multicolumn{2}{c}{\textbf{HPSv2.1}} &
  \multicolumn{2}{c}{\textbf{PickScore}} &
  \multicolumn{2}{c}{\textbf{ImageReward}} &
  \multicolumn{2}{c}{\textbf{UR-Align}} &
  \multicolumn{2}{c}{\textbf{UR-IQ}} \\
  \cmidrule(lr){2-3} \cmidrule(lr){4-5} \cmidrule(lr){6-7} \cmidrule(lr){8-9} \cmidrule(lr){10-11}
  & Score$\uparrow$ & \dt{$\Delta$($\uparrow$)}
  & Score$\uparrow$ & \dt{$\Delta$($\uparrow$)}
  & Score$\uparrow$ & \dt{$\Delta$($\uparrow$)}
  & Score$\uparrow$ & \dt{$\Delta$($\uparrow$)}
  & Score$\uparrow$ & \dt{$\Delta$($\uparrow$)} \\
  \midrule
  \multicolumn{11}{l}{\colorbox{groupShade}{\textbf{\textit{SD3.5-M}}}} \\
  \midrule
  \texttt{SD3.5-M}
  & 0.2881 & \dt{---}
  & 22.5004 & \dt{---}
  & 0.9504 & \dt{---}
  & 3.3935 & \dt{---}
  & 3.6376 & \dt{---} \\
  ReFL
  & 0.3872 & \dt{$+$0.0991}
  & 23.4021 & \dt{$+$0.9017}
  & 1.3799 & \dt{$+$0.4295}
  & 3.4779 & \dt{$+$0.0844}
  & 3.9464 & \dt{$+$0.3088} \\
  DRaFT-LV
  & 0.3818 & \dt{$+$0.0937}
  & 23.3506 & \dt{$+$0.8502}
  & 1.3673 & \dt{$+$0.4169}
  & \second{3.5072} & \dt{$\second{+0.1137}$}
  & 3.9245 & \dt{$+$0.2869} \\
  DRTune
  & 0.3870 & \dt{$+$0.0989}
  & 23.2882 & \dt{$+$0.7878}
  & 1.3794 & \dt{$+$0.4290}
  & 3.4562 & \dt{$+$0.0627}
  & 3.8869 & \dt{$+$0.2493} \\
  LeapAlign
  & 0.3924 & \dt{$+$0.1043}
  & 23.3142 & \dt{$+$0.8138}
  & \second{1.4425} & \dt{$\second{+0.4921}$}
  & 3.4854 & \dt{$+$0.0919}
  & 3.9030 & \dt{$+$0.2654} \\
  \midrule
  \rowcolor{colSky!15}
  \ourscompose
  & 0.3939 & \dt{$+$0.1058}
  & 23.4104 & \dt{$+$0.9100}
  & 1.4299 & \dt{$+$0.4795}
  & 3.4774 & \dt{$+$0.0839}
  & \best{3.9690} & \dt{$\mathbf{+0.3314}$} \\
  \rowcolor{colSky!20}
  \ourscomposeleap
  & \second{0.3966} & \dt{$\second{+0.1085}$}
  & \second{23.4130} & \dt{$\second{+0.9126}$}
  & 1.4386 & \dt{$+$0.4882}
  & 3.4712 & \dt{$+$0.0777}
  & 3.9481 & \dt{$+$0.3105} \\
  \rowcolor{colSky!30}
  \oursleap
  & \best{0.3998} & \dt{$\mathbf{+0.1117}$}
  & \best{23.5040} & \dt{$\mathbf{+1.0036}$}
  & \best{1.4538} & \dt{$\mathbf{+0.5034}$}
  & \best{3.5127} & \dt{$\mathbf{+0.1192}$}
  & \second{3.9680} & \dt{$\second{+0.3304}$} \\
  \midrule
  \multicolumn{11}{l}{\colorbox{groupShade}{\textbf{\textit{FLUX.1-dev 12B}}}} \\
  \midrule
  \texttt{FLUX.1-dev}
  & 0.3016 & \dt{---}
  & 22.5940 & \dt{---}
  & 1.0094 & \dt{---}
  & 3.3168 & \dt{---}
  & 3.7551 & \dt{---} \\
  ReFL
  & 0.3930 & \dt{$+$0.0914}
  & 23.6886 & \dt{$+$1.0946}
  & 1.3916 & \dt{$+$0.3822}
  & 3.5005 & \dt{$+$0.1837}
  & 4.0293 & \dt{$+$0.2742} \\
  DRaFT-LV
  & 0.3900 & \dt{$+$0.0884}
  & 23.6290 & \dt{$+$1.0350}
  & 1.3910 & \dt{$+$0.3816}
  & 3.5264 & \dt{$+$0.2096}
  & 4.0259 & \dt{$+$0.2708} \\
  DRTune
  & 0.3920 & \dt{$+$0.0904}
  & 23.6704 & \dt{$+$1.0764}
  & 1.4273 & \dt{$+$0.4179}
  & 3.5403 & \dt{$+$0.2235}
  & 4.0697 & \dt{$+$0.3146} \\
  LeapAlign
  & 0.4063 & \dt{$+$0.1047}
  & 23.6158 & \dt{$+$1.0218}
  & 1.4373 & \dt{$+$0.4279}
  & 3.5290 & \dt{$+$0.2122}
  & 4.0584 & \dt{$+$0.3033} \\
  \midrule
  \rowcolor{colSky!15}
  \ourscompose
  & 0.4081 & \dt{$+$0.1065}
  & 23.7458 & \dt{$+$1.1518}
  & 1.4742 & \dt{$+$0.4648}
  & 3.5408 & \dt{$+$0.2240}
  & 4.0897 & \dt{$+$0.3346} \\
  \rowcolor{colSky!20}
  \ourscomposeleap
  & \second{0.4119} & \dt{$\second{+0.1103}$}
  & \best{23.7952} & \dt{$\mathbf{+1.2012}$}
  & \best{1.5156} & \dt{$\mathbf{+0.5062}$}
  & \second{3.5681} & \dt{$\second{+0.2513}$}
  & \best{4.1098} & \dt{$\mathbf{+0.3547}$} \\
  \rowcolor{colSky!30}
  \oursleap
  & \best{0.4134} & \dt{$\mathbf{+0.1118}$}
  & \second{23.7796} & \dt{$\second{+1.1856}$}
  & \second{1.4906} & \dt{$\second{+0.4812}$}
  & \best{3.5848} & \dt{$\mathbf{+0.2680}$}
  & \second{4.0950} & \dt{$\second{+0.3399}$} \\
  \midrule
  \multicolumn{11}{l}{\colorbox{groupShade}{\textbf{\textit{FLUX.2-Klein-base 9B}}}} \\
  \midrule
  \texttt{FLUX.2-Klein-base}
  & 0.2873 & \dt{---}
  & 22.3158 & \dt{---}
  & 1.1405 & \dt{---}
  & 3.6392 & \dt{---}
  & 3.7982 & \dt{---} \\
  ReFL
  & 0.4272 & \dt{$+$0.1399}
  & 23.6002 & \dt{$+$1.2844}
  & 1.6245 & \dt{$+$0.4840}
  & \second{3.6817} & \dt{$\second{+0.0425}$}
  & 4.0565 & \dt{$+$0.2583} \\
  DRaFT-LV
  & 0.4233 & \dt{$+$0.1360}
  & 23.6288 & \dt{$+$1.3130}
  & 1.6262 & \dt{$+$0.4857}
  & 3.6495 & \dt{$+$0.0103}
  & 3.9319 & \dt{$+$0.1337} \\
  DRTune
  & 0.4122 & \dt{$+$0.1249}
  & 23.5716 & \dt{$+$1.2558}
  & 1.6264 & \dt{$+$0.4859}
  & 3.6638 & \dt{$+$0.0246}
  & 4.0589 & \dt{$+$0.2607} \\
  LeapAlign
  & 0.4246 & \dt{$+$0.1373}
  & 23.4884 & \dt{$+$1.1726}
  & \second{1.6553} & \dt{$\second{+0.5148}$}
  & \best{3.6926} & \dt{$\mathbf{+0.0534}$}
  & 4.0625 & \dt{$+$0.2643} \\
  \midrule
  \rowcolor{colSky!15}
  \ourscompose
  & \second{0.4384} & \dt{$\second{+0.1511}$}
  & \best{23.8446} & \dt{$\mathbf{+1.5288}$}
  & \best{1.6649} & \dt{$\mathbf{+0.5244}$}
  & 3.6720 & \dt{$+$0.0328}
  & \best{4.0787} & \dt{$\mathbf{+0.2805}$} \\
  \rowcolor{colSky!20}
  \ourscomposeleap
  & \best{0.4387} & \dt{$\mathbf{+0.1514}$}
  & \second{23.7718} & \dt{$\second{+1.4560}$}
  & 1.6491 & \dt{$+$0.5086}
  & 3.6775 & \dt{$+$0.0383}
  & \second{4.0720} & \dt{$\second{+0.2738}$} \\
  \rowcolor{colSky!30}
  \oursleap
  & 0.4321 & \dt{$+$0.1448}
  & 23.6938 & \dt{$+$1.3780}
  & 1.6376 & \dt{$+$0.4971}
  & 3.6705 & \dt{$+$0.0313}
  & 4.0652 & \dt{$+$0.2670} \\
  \bottomrule
  \end{tabular}
}
\end{table}

%% file: fig/Flux1_image_comparsion.tex
\begin{figure}[t]
    \centering
    \includegraphics[width=1.00\textwidth]{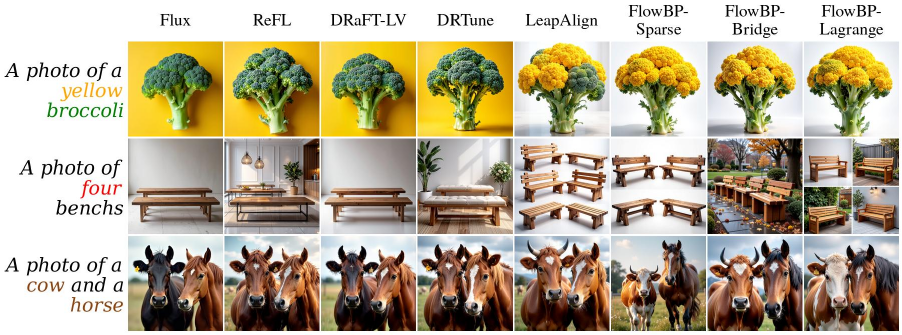}
    \vspace{-12pt}
    \caption{
    Qualitative comparison on GenEval prompts with \texttt{FLUX.1-dev}. Each row corresponds to one compositional prompt, and columns are ordered from left to right by method. The examples compare visual fidelity and prompt following under object, attribute, counting, and spatial constraints.
    }
    \label{fig:flux1_geneval_qualitative}
\end{figure}

%% file: tab/geneval.tex
\providecommand{\best}[1]{\textbf{#1}}
\providecommand{\second}[1]{\underline{#1}}
\definecolor{deltaGray}{gray}{0.45}
\definecolor{colSky}{RGB}{135,206,235}
\definecolor{groupShade}{RGB}{238,238,238}
\providecommand{\dt}[1]{}
\renewcommand{\dt}[1]{\textcolor{deltaGray}{\scriptsize #1}}

\begin{table}[!htbp]
\centering
\caption{
GenEval results on FLUX.1-dev, including the overall score and task-level accuracies. $\Delta$ denotes the improvement over the base model. Best and second-best results are shown in \textbf{bold} and \underline{underlined}, respectively. Blue rows indicate our methods.
}
\label{tab:geneval_flux1}
\setlength{\tabcolsep}{1.9pt}
\renewcommand{\arraystretch}{1.12}
\resizebox{0.95\textwidth}{!}{
  \begin{tabular}{@{}l cc cc cc cc cc cc cc@{}}
  \toprule
  \multirow{2}{*}{\textbf{Method}} &
  \multicolumn{2}{c}{\textbf{Overall}} &
  \multicolumn{2}{c}{\textbf{Single Obj.}} &
  \multicolumn{2}{c}{\textbf{Two Obj.}} &
  \multicolumn{2}{c}{\textbf{Count}} &
  \multicolumn{2}{c}{\textbf{Color}} &
  \multicolumn{2}{c}{\textbf{Pos}} &
  \multicolumn{2}{c}{\textbf{AttrB}} \\
  \cmidrule(lr){2-3} \cmidrule(lr){4-5} \cmidrule(lr){6-7}
  \cmidrule(lr){8-9} \cmidrule(lr){10-11} \cmidrule(lr){12-13}
  \cmidrule(l){14-15}
  & Score$\uparrow$ & \dt{$\Delta$($\uparrow$)}
  & Score$\uparrow$ & \dt{$\Delta$($\uparrow$)}
  & Score$\uparrow$ & \dt{$\Delta$($\uparrow$)}
  & Score$\uparrow$ & \dt{$\Delta$($\uparrow$)}
  & Score$\uparrow$ & \dt{$\Delta$($\uparrow$)}
  & Score$\uparrow$ & \dt{$\Delta$($\uparrow$)}
  & Score$\uparrow$ & \dt{$\Delta$($\uparrow$)} \\
  \midrule
  \multicolumn{15}{l}{\colorbox{groupShade}{\textbf{\textit{FLUX.1-dev 12B}}}} \\
  \midrule
  \texttt{FLUX.1-dev}
  & 63.25 & \dt{---}
  & 97.81 & \dt{---}
  & 76.52 & \dt{---}
  & 70.31 & \dt{---}
  & 76.86 & \dt{---}
  & 17.75 & \dt{---}
  & 40.25 & \dt{---} \\
  ReFL
  & 65.94 & \dt{$+$2.69}
  & 97.50 & \dt{$-$0.31}
  & 77.27 & \dt{$+$0.75}
  & 71.88 & \dt{$+$1.57}
  & \second{78.99} & \dt{$\second{+2.13}$}
  & 22.25 & \dt{$+$4.50}
  & 47.75 & \dt{$+$7.50} \\
  DRaFT-LV
  & 68.31 & \dt{$+$5.06}
  & 98.13 & \dt{$+$0.32}
  & \best{86.36} & \dt{$\mathbf{+9.84}$}
  & \best{75.00} & \dt{$\mathbf{+4.69}$}
  & 77.13 & \dt{$+$0.27}
  & 23.25 & \dt{$+$5.50}
  & 50.00 & \dt{$+$9.75} \\
  DRTune
  & 66.78 & \dt{$+$3.53}
  & \second{98.75} & \dt{$\second{+0.94}$}
  & 80.56 & \dt{$+$4.04}
  & 72.81 & \dt{$+$2.50}
  & 76.33 & \dt{$-$0.53}
  & 20.75 & \dt{$+$3.00}
  & 51.50 & \dt{$+$11.25} \\
  LeapAlign
  & 67.44 & \dt{$+$4.19}
  & 98.44 & \dt{$+$0.63}
  & 84.09 & \dt{$+$7.57}
  & 72.19 & \dt{$+$1.88}
  & 78.19 & \dt{$+$1.33}
  & 22.25 & \dt{$+$4.50}
  & 49.50 & \dt{$+$9.25} \\
  \midrule
  \rowcolor{colSky!15}
  \ourscompose
  & 68.19 & \dt{$+$4.94}
  & \second{98.75} & \dt{$\second{+0.94}$}
  & 84.85 & \dt{$+$8.33}
  & \second{73.13} & \dt{$\second{+2.82}$}
  & 78.19 & \dt{$+$1.33}
  & 22.00 & \dt{$+$4.25}
  & 52.25 & \dt{$+$12.00} \\
  \rowcolor{colSky!20}
  \ourscomposeleap
  & \second{68.75} & \dt{$\second{+5.50}$}
  & 96.88 & \dt{$-$0.93}
  & \second{85.10} & \dt{$\second{+8.58}$}
  & 70.94 & \dt{$+$0.63}
  & 76.86 & \dt{$+$0.00}
  & \best{27.25} & \dt{$\mathbf{+9.50}$}
  & \second{55.50} & \dt{$\second{+15.25}$} \\
  \rowcolor{colSky!30}
  \oursleap
  & \best{69.88} & \dt{$\mathbf{+6.63}$}
  & \best{99.06} & \dt{$\mathbf{+1.25}$}
  & \second{85.10} & \dt{$\second{+8.58}$}
  & 71.88 & \dt{$+$1.57}
  & \best{79.52} & \dt{$\mathbf{+2.66}$}
  & \second{23.75} & \dt{$\second{+6.00}$}
  & \best{60.00} & \dt{$\mathbf{+19.75}$} \\
  \bottomrule
  \end{tabular}
}
\end{table}

%% file: fig/Flux1_highorder_ablation.tex
\begin{figure}[t]
    \centering
\includegraphics[width=1\textwidth]{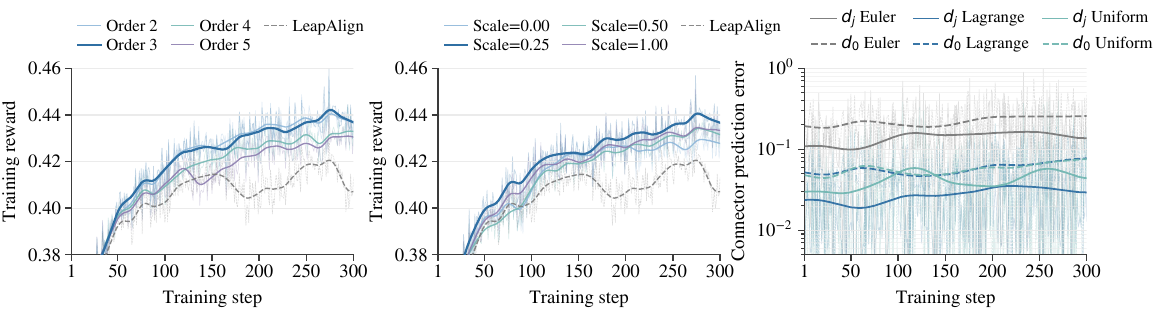}
\vspace{-12pt}    
\caption{Ablation of \oursleap on \texttt{FLUX.1-dev}. Left: effect of the Lagrange quadrature order on HPSv2.1 training reward. Middle: effect of the gradient-support scale. Right: connector prediction errors for Euler, Lagrange, and uniform connectors, where uniform uses the same supports as Lagrange but replaces the integrated coefficients with equal weights.
}  \label{fig:flux1_highorder_ablation}
\end{figure}

%% file: fig/ComposeLeap_ablation.tex
\begin{figure}[t]
    \centering
    \begin{minipage}[t]{0.62\textwidth}
        \vspace{0pt}
        \centering
        \vbox to 4.2cm{\vfil\hbox to \textwidth{\hfil\includegraphics[width=\textwidth,height=4cm,keepaspectratio]{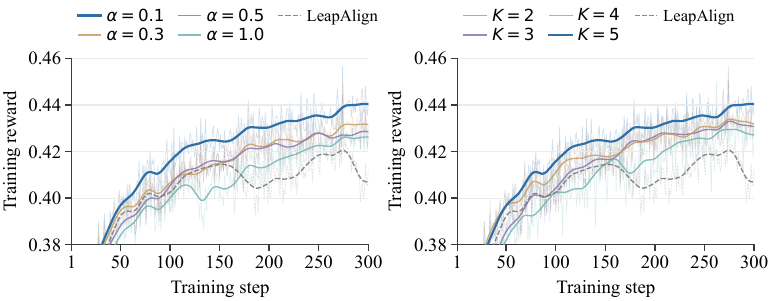}\hfil}\vfil}

        \caption{Ablation of \ourscomposeleap. Left: effect of the nested-gradient scale. Right: effect of active-step budget $K$.}
        \label{fig:composeleap_ablation}
    \end{minipage}
    \hfill
    \begin{minipage}[t]{0.35\textwidth}
        \vspace{0pt}
        \centering
        \vbox to 4.2cm{\vfil\hbox to \textwidth{\hfil\includegraphics[width=\textwidth,height=4cm,keepaspectratio]{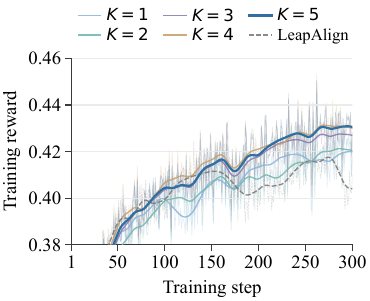}\hfil}\vfil}

        \caption{Ablation of \ourscompose{} active-step budget $K$.}
        \label{fig:compose_ablation}
    \end{minipage}
\end{figure}

%% file: fig/Loss_d0_comparsion.tex
\begin{figure}[t]
    \centering
    \includegraphics[width=\textwidth]{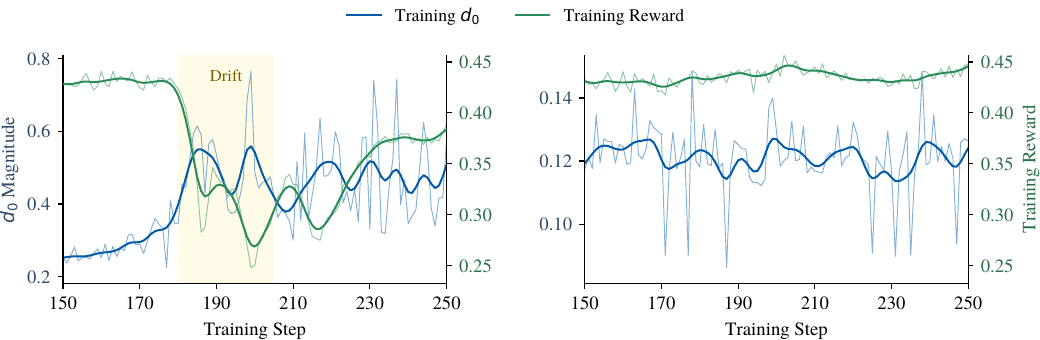}
    \vspace{-12pt}
    \caption{
    Connector residual and reward dynamics during training. Left: in LeapAlign, we observe that excessively large endpoint residual $d_0$ can coincide with reward collapse, as highlighted by the shaded interval. Right: \oursleap keeps $d_0$ smaller through the high-order connector, making this failure mode less likely during training.
    }
    \label{fig:d0_reward_dynamics}
\end{figure}

%% file: A-tab/endpoint_connector_ablation.tex
\providecommand{\best}[1]{\textbf{#1}}
\providecommand{\second}[1]{\underline{#1}}
\providecommand{\dt}[1]{}
\definecolor{deltaGray}{gray}{0.45}
\renewcommand{\dt}[1]{\textcolor{deltaGray}{\scriptsize #1}}
\definecolor{colSky}{RGB}{135,206,235}
\definecolor{groupShade}{RGB}{238,238,238}

\begin{table}[t]
\centering
\caption{
Effect of \emph{endpoint reconstruction} on \texttt{FLUX.1-dev}. ReFL and DRTune evaluate the reward on the posterior-mean estimate $\hat\vx_0 = \E[\vx_0\mid\vx_i]$ by default. We replace this with the same Euler-reconstructed endpoint treatment used by \ourscompose{} (``$+$ endpoint recon.''), which replays the cached rollout with selected active velocities re-forwarded so the reward is evaluated on the cached sampled image $\vx_0$ by construction. $\Delta$ denotes the empirical change over the same method without endpoint reconstruction. Blue rows use endpoint reconstruction.
}
\label{tab:endpoint_reconstruction}
\vspace{3pt}
\scriptsize
\setlength{\tabcolsep}{1.7pt}
\renewcommand{\arraystretch}{1.13}
\resizebox{1.0\textwidth}{!}{
  \begin{tabular}{@{}l cc cc cc cc cc@{}}
  \toprule
  \multirow{3}{*}{\textbf{Method}} &
  \multicolumn{2}{c}{\textbf{In-Domain}} &
  \multicolumn{8}{c}{\textbf{Out-of-Domain}} \\
  \cmidrule(lr){2-3} \cmidrule(lr){4-11}
  &
  \multicolumn{2}{c}{\textbf{HPSv2.1}} &
  \multicolumn{2}{c}{\textbf{PickScore}} &
  \multicolumn{2}{c}{\textbf{ImageReward}} &
  \multicolumn{2}{c}{\textbf{UR-Align}} &
  \multicolumn{2}{c}{\textbf{UR-IQ}} \\
  \cmidrule(lr){2-3} \cmidrule(lr){4-5} \cmidrule(lr){6-7} \cmidrule(lr){8-9} \cmidrule(lr){10-11}
  & Score$\uparrow$ & \dt{$\Delta$($\uparrow$)}
  & Score$\uparrow$ & \dt{$\Delta$($\uparrow$)}
  & Score$\uparrow$ & \dt{$\Delta$($\uparrow$)}
  & Score$\uparrow$ & \dt{$\Delta$($\uparrow$)}
  & Score$\uparrow$ & \dt{$\Delta$($\uparrow$)} \\
  \midrule
  \multicolumn{11}{l}{\colorbox{groupShade}{\textbf{\textit{FLUX.1-dev 12B}}}} \\
  \midrule
  ReFL
  & 0.3930 & \dt{---}
  & 23.6886 & \dt{---}
  & 1.3916 & \dt{---}
  & 3.5005 & \dt{---}
  & 4.0293 & \dt{---} \\
  \rowcolor{colSky!20}
  \quad $+$ endpoint recon.
  & \best{0.3993} & \dt{$+$0.0063}
  & \best{23.8121} & \dt{$+$0.1235}
  & \best{1.4117} & \dt{$+$0.0201}
  & \best{3.5272} & \dt{$+$0.0267}
  & \best{4.0532} & \dt{$+$0.0239} \\
  \midrule
  DRTune
  & 0.3920 & \dt{---}
  & 23.6704 & \dt{---}
  & 1.4273 & \dt{---}
  & 3.5403 & \dt{---}
  & 4.0697 & \dt{---} \\
  \rowcolor{colSky!20}
  \quad $+$ endpoint recon.
  & \best{0.4004} & \dt{$+$0.0085}
  & \best{23.7039} & \dt{$+$0.0335}
  & \best{1.4576} & \dt{$+$0.0303}
  & \best{3.5645} & \dt{$+$0.0242}
  & \best{4.0816} & \dt{$+$0.0119} \\
  \bottomrule
  \end{tabular}
}
\end{table}

%% file: sec/05_conclusion.tex
\section{Conclusion}
\label{sec:conclusion}

We presented \ours{}, a unified surrogate-trajectory framework for direct reward backpropagation in text-to-image flow matching models. By separating reward-model input, active-set selection, integration weights, and bridge coupling, the framework recovers prior methods as special cases while exposing previously unexplored regions of the design space. The resulting instantiations, \ourscompose{}, \ourscomposeleap{}, and \oursleap{}, bound memory by the number of active velocities and avoid long Jacobian chains by construction. Across three backbones and multiple preference, quality, and compositional metrics, these variants consistently improve over direct-gradient baselines, with complementary strengths across models and reward dimensions. More importantly, the results show that the four design axes provide actionable control over the surrogate backward trajectory: endpoint reconstruction improves the reward-model input, sparse active sets trade compute for trajectory coverage, higher-order integration improves long-interval accuracy, and bridge coupling adds bounded cross-step credit assignment when useful. These findings suggest that explicit surrogate-trajectory design is a practical and interpretable route for scaling reward alignment in flow-based generative models.

%% file: A-sec/01_derivations.tex
\providecommand{\sg}{\operatorname{sg}}
\providecommand{\A}{\mathcal{A}}
\providecommand{\Apre}{\mathcal{A}_{\mathrm{pre}}}
\providecommand{\Apost}{\mathcal{A}_{\mathrm{post}}}
\providecommand{\dvt}[1]{\frac{\partial \vv_\theta(\vx_{#1},\sigma_{#1},\vc)}{\partial \theta}}
\providecommand{\dvx}[1]{\frac{\partial \vv_\theta(\vx_{#1},\sigma_{#1},\vc)}{\partial \vx_{#1}}}
\providecommand{\dvtildex}[1]{\frac{\partial \vv_\theta(\vx_{#1},\sigma_{#1},\vc)}{\partial \theta}}
\providecommand{\dvxtildex}[1]{\frac{\partial \vv_\theta(\vx_{#1},\sigma_{#1},\vc)}{\partial \vx_{#1}}}

\section{Additional Derivations}
\label{app:additional_derivations}

This section gives the per-method derivations of the unified gradient \cref{eq:unified_grad}. We follow the notation of \cref{sec:method}: the cached no-gradient rollout produces detached states and velocities $\{(\vx_i,\vv_i,\sigma_i)\}_{i=0}^{N}$ satisfying $\vx_{i-1}=\vx_i-h_i\vv_i$ with $h_i=\sigma_i-\sigma_{i-1}>0$ (cf.\ \cref{eq:euler}). The surrogate replaces each cached velocity with $\tilde\vv_i$ from \cref{eq:surrogate_v}, parameterized by an active set $\A=\Apre\cup\Apost$, a bridge index $j$ with $0\le j<k$, and a nested-gradient scale $\alpha\in[0,1]$. All cached states are treated as constants during the second forward graph. Connector-based surrogates use the straight-through connector in \cref{eq:connector}; reconstruction-based surrogates instead replay the cached Euler interval with $\tilde\vv_i$, so their forward states equal the cached rollout states by construction.

\subsection{Surrogate-Fidelity Metrics}
\label{app:surrogate_fidelity}
For the diagnostic comparison in \cref{sec:method}, we summarize each surrogate along two axes. Reward-path fidelity measures whether the active steps cover enough trajectory weight and whether their endpoint estimates match the actual generated sample:
\begin{equation}
\label{eq:reward_path_fidelity}
\RewardPathFidelity.
\end{equation}
Here $\xhat{i}$ is the endpoint estimate induced by active step $i$, $w_i$ is its surrogate integration weight, $h_i$ is the corresponding full Euler weight, and $\A$ is the active set. A large $\mathcal{C}$ therefore requires both accurate endpoint reconstruction and sufficient rollout-weight coverage. Nested-coupling fidelity instead measures whether the surrogate retains an accurate gradient path across denoising steps:
\begin{equation}
\label{eq:nested_coupling_fidelity}
\LeapQuality , \qquad \NestedCouplingFidelity .
\end{equation}
Here $\M$ is the surrogate backward graph, and $\NestedDepth$ counts the maximum number of inter-step Jacobian factors on any gradient path. Thus full backpropagation has $\NestedDepth=N-1$, detached surrogates have $\NestedDepth=0$, and bridged surrogates have $\NestedDepth=1$. The argument $\M$ simply makes explicit which surrogate graph is being measured. Thus $\mathcal{C}$ describes the forward sample consumed by the reward model, whereas $\Fnest$ describes the backward coupling carried by the gradient.

\subsection{\oursleap}
\label{app:high_order_leapalign}

\oursleap{} keeps the two-segment structure of LeapAlign but replaces each one-point leap by an integrated polynomial approximation of the velocity field. Let $k>j>0$ be the sampled bridge indices. The quadrature supports are
\begin{equation}
    \mathcal{S}_{\mathrm{pre}}\subseteq\{j+1,\ldots,k\},
    \qquad k\in\mathcal{S}_{\mathrm{pre}},
    \label{eq:app_ho_support_pre}
\end{equation}
and analogously $\mathcal{S}_{\mathrm{post}}\subseteq\{1,\ldots,j\}$ with anchor $j\in\mathcal{S}_{\mathrm{post}}$. The gradient-active sets satisfy $\Apre\subseteq\mathcal{S}_{\mathrm{pre}}$ and $\Apost\subseteq\mathcal{S}_{\mathrm{post}}$, with $K=|\Apre\cup\Apost|\le M$. Inactive supports are included only in the forward quadrature through detached cached velocities, while non-anchor active supports are attenuated during backpropagation. The pre-segment Lagrange basis polynomial associated with support $s_m\in\mathcal{S}_{\mathrm{pre}}$ is
\begin{equation}
    \ell_m^{\mathcal{S}_{\mathrm{pre}}}(\sigma)
    =
    \prod_{n\neq m}
    \frac{\sigma-\sigma_{s_n}}{\sigma_{s_m}-\sigma_{s_n}},
    \label{eq:app_ho_lagrange_basis}
\end{equation}
with integrated quadrature weight
\begin{equation}
    w_{s_m}^{\mathrm{L}}
    =
    \int_{\sigma_j}^{\sigma_k} \ell_m^{\mathcal{S}_{\mathrm{pre}}}(\sigma)\,d\sigma .
    \label{eq:app_ho_lagrange_weight}
\end{equation}
Post-segment quantities $\ell_m^{\mathcal{S}_{\mathrm{post}}}$ and $w_{s'_m}^{\mathrm{L}}$ are constructed analogously over $[\sigma_0,\sigma_j]$. With $\sigma_j<\sigma_k$ and $\sigma_0<\sigma_j$, the Lagrange weights are non-negative for the small per-segment support counts we use (in particular three supports per segment, which gives Simpson-like positive weights) and play the role of the positive Euler step $h_i$ in the surrogate; we write $w_i^{\mathrm{L}}$ without segment subscript when the segment is clear from $i$ ($\mathcal{S}_{\mathrm{pre}}$ and $\mathcal{S}_{\mathrm{post}}$ are disjoint).

The surrogate velocity $\tilde\vv_i$ at any active support $i\in\A$ is given by \cref{eq:surrogate_v}: the post-segment anchor $i=j$ uses the bridge case $\vv_\theta\bigl(\alpha\,\vx_j+(1-\alpha)\,\sg(\vx_j),\,\sigma_j,\,\vc\bigr)$, and every other active support uses the regular active case $\vv_\theta(\sg(\vx_i),\sigma_i,\vc)$. Inactive supports use detached cached velocities in the forward quadrature. To prevent the high-order correction from inflating the gradient norm, we attenuate the contribution of non-anchor active supports through the stop-gradient identity
\begin{equation}
    \tilde\vv_i^{\rho}
    =
    \sg(\tilde\vv_i)+\rho_i\bigl(\tilde\vv_i-\sg(\tilde\vv_i)\bigr),
    \qquad \rho_i\in[0,1],
    \label{eq:app_ho_support_scale}
\end{equation}
with $\rho_i=1$ for the anchor supports $i\in\{k,j\}$ and $\rho_i=g_s$ for non-anchor active supports. For inactive supports $i\notin\A$, we set $\tilde\vv_i^{\rho}=\sg(\vv_i)$, so they contribute to the forward quadrature but not to the backward graph. The forward value is unchanged; the backward signal through active support $i$ is scaled by $\rho_i$.

The pre-segment surrogate and bridge connector are
\begin{align}
    \hat\vx_j
    &=
    \vx_k-
    \sum_{i\in\mathcal{S}_{\mathrm{pre}}}
    w_i^{\mathrm{L}}\,\tilde\vv_i^{\rho},
    \label{eq:app_ho_pred_j}
    \\
    \vx_j
    &=
    \hat\vx_j+\sg(\vx_j-\hat\vx_j).
    \label{eq:app_ho_conn_j}
\end{align}
The just-pinned $\vx_j$ is the latent that enters the bridge case of \cref{eq:surrogate_v} when computing $\tilde\vv_j$. The post-segment surrogate and endpoint connector are
\begin{align}
    \hat\vx_0
    &=
    \vx_j-
    \sum_{i\in\mathcal{S}_{\mathrm{post}}}
    w_i^{\mathrm{L}}\,\tilde\vv_i^{\rho},
    \label{eq:app_ho_pred_0}
    \\
    \vx_0
    &=
    \hat\vx_0+\sg(\vx_0-\hat\vx_0).
    \label{eq:app_ho_conn_0}
\end{align}

Because the bridge connector \cref{eq:app_ho_conn_j} is transparent to gradients,
\begin{equation}
    \frac{\partial \vx_j}{\partial\theta}
    =
    -\sum_{i\in\Apre}
    \rho_i\,w_i^{\mathrm{L}}\,
    \dvt{i}.
    \label{eq:app_ho_grad_pre}
\end{equation}
Applying the chain rule at the bridge ($i=j$, $\rho_j=1$) gives
\begin{align}
    \frac{\partial \vx_0}{\partial\theta}
    &=
    \frac{\partial \vx_j}{\partial\theta}
    -
    \sum_{i\in\Apost\setminus\{j\}}
    \rho_i\,w_i^{\mathrm{L}}\,\dvt{i}
    -
    w_j^{\mathrm{L}}\,\dvtildex{j}
    -\,
    \alpha\,w_j^{\mathrm{L}}\,\dvxtildex{j}\,
    \frac{\partial \vx_j}{\partial\theta} .
    \label{eq:app_ho_grad_full}
\end{align}
Substituting \cref{eq:app_ho_grad_pre} and combining direct terms,
\begin{align}
    \frac{\partial \vx_0}{\partial\theta}
    &=
    -\sum_{i\in\A}
    \rho_i\,w_i^{\mathrm{L}}\,\dvt{i}
    \;+\;
    \alpha\,w_j^{\mathrm{L}}\,\dvxtildex{j}
    \sum_{i\in\Apre}
    \rho_i\,w_i^{\mathrm{L}}\,\dvt{i}.
    \label{eq:app_ho_grad_collected}
\end{align}
The last term is the only nested term: it contains exactly one Jacobian factor and is scaled by $\alpha$. With a single support per segment (only the segment-start anchors $k$ and $j$), $w_k^{\mathrm{L}}=\sigma_k-\sigma_j$ and $w_j^{\mathrm{L}}=\sigma_j$, recovering the LeapAlign gradient.

\subsection{\ourscompose}
\label{app:composed_velocity}

\ourscompose{} drops the bridge ($j=0$, $\alpha=0$, $\Apost=\emptyset$) and reconstructs the endpoint by replaying the cached trajectory with Euler weights. With no bridge, the bridge case of \cref{eq:surrogate_v} does not arise: every $i\in\A=\Apre$ uses the regular active form $\vv_\theta(\sg(\vx_i),\sigma_i,\vc)$, and every $i\notin\A$ uses $\sg(\vv_i)$. The endpoint reconstruction is
\begin{equation}
    \vx_0
    =
    \vx_N-
    \sum_{i=1}^{N} h_i\,\tilde\vv_i.
    \label{eq:app_compose_endpoint}
\end{equation}
Inactive velocities are detached and contribute zero to the backward pass; their forward values telescope with the active terms to match the cached $\vx_0$ exactly, so no endpoint connector is needed. Since every active velocity is evaluated on a detached input, no inter-step Jacobian appears, and
\begin{equation}
    \frac{\partial \vx_0}{\partial\theta}
    =
    -\sum_{i\in\A} h_i\,\dvt{i}.
    \label{eq:app_compose_grad}
\end{equation}
\ourscompose{} thus removes the nested gradient path in favor of a dense but fully decoupled set of direct terms.

\subsection{\ourscomposeleap}
\label{app:compose_split}

\ourscomposeleap{} adds a single bridge to \ourscompose{}'s full Euler-quadrature trajectory. The split index $j\in\{1,\ldots,N-1\}$ partitions the rollout into a pre-segment $(N\to j)$ and a post-segment $(j\to0)$. The pre-segment active set $\Apre\subseteq\{j+1,\ldots,N\}$ and the post-segment active set $\Apost\subseteq\{1,\ldots,j\}$ jointly form $\A=\Apre\cup\Apost$, with $j\in\Apost$ as the bridge anchor. All weights are Euler weights $w_i=h_i$. Each $\tilde\vv_i$ is given by \cref{eq:surrogate_v}: regular active form for $i\in\Apre$ and $i\in\Apost\setminus\{j\}$, and the bridge case at $i=j$.

The pre-segment reconstruction is
\begin{equation}
    \vx_j
    =
    \vx_N-
    \sum_{i=j+1}^{N} h_i\,\tilde\vv_i.
    \label{eq:app_cs_recon_j}
\end{equation}
The post-segment endpoint reconstruction is
\begin{equation}
    \vx_0
    =
    \vx_j-
    \sum_{i=1}^{j} h_i\,\tilde\vv_i.
    \label{eq:app_cs_recon_0}
\end{equation}
Both reconstructions include inactive cached velocities, whose forward values make the reconstructed $\vx_j$ and $\vx_0$ match the cached rollout states exactly. Since inactive velocities are detached,
\begin{equation}
    \frac{\partial \vx_j}{\partial\theta}
    =
    -\sum_{i\in\Apre} h_i\,\dvt{i}.
    \label{eq:app_cs_grad_pre}
\end{equation}
At the bridge ($i=j$), the chain rule contributes both a direct term and a nested term through $\vx_j$:
\begin{align}
    \frac{\partial \vx_0}{\partial\theta}
    &=
    \frac{\partial \vx_j}{\partial\theta}
    -
    \sum_{i\in\Apost\setminus\{j\}} h_i\,\dvt{i}
    -
    h_j\,\dvtildex{j}
    -\,
    \alpha\,h_j\,\dvxtildex{j}\,
    \frac{\partial \vx_j}{\partial\theta}.
\end{align}
Substituting \cref{eq:app_cs_grad_pre} and combining direct terms,
\begin{equation}
    \frac{\partial \vx_0}{\partial\theta}
    =
    -\sum_{i\in\A} h_i\,\dvt{i}
    \;+\;
    \alpha\,h_j\,\dvxtildex{j}
    \sum_{i\in\Apre} h_i\,\dvt{i}.
    \label{eq:app_cs_grad_full}
\end{equation}
This expression matches the unified gradient \cref{eq:unified_grad}: the first sum is the direct active-step update over both segments, and the last term is the bridge-induced nested term scaled by $\alpha$. Setting $\alpha=0$ removes the nested gradient path; $\alpha=1$ preserves the full one-Jacobian bridge dependence.

%% file: A-sec/02_experimental_details.tex
\section{Additional Experimental Details}
\label{appsec:experimental_details}

\noindent\textbf{Training configuration.}
\cref{tab:app_training_config} summarizes the common training configuration used in our experiments. All methods are trained with HPSv2.1 as the differentiable reward model and use the hinge reward loss
\begin{equation}
    \mathcal{L}_{\mathrm{reward}} = \mathrm{ReLU}(\lambda - r(\vx_0, \vc)),
\end{equation}
where $\vx_0$ is the predicted clean image, $\vc$ is the text prompt, and $\lambda=0.55$. Following the LeapAlign setup, connector-based methods further apply trajectory-similarity weighting to this raw hinge loss. For a connector-based bridged surrogate, we measure the connector residuals at the bridge and clean endpoint,
\begin{equation}
    d_j = \|\vx_j-\hat\vx_j\|_2,
    \qquad
    d_0 = \|\vx_0-\hat\vx_0\|_2,
\end{equation}
where $\hat\vx_j$ and $\hat\vx_0$ are the surrogate predictions before the straight-through connectors. We then form
\begin{equation}
    w_{\mathrm{sim}}
    =
    \frac{1}{\max(d_j,\tau)+\max(d_0,\tau)},
    \qquad
    \mathcal{L}
    =
    \sg(w_{\mathrm{sim}})\,\mathcal{L}_{\mathrm{reward}} .
\end{equation}
Here $\tau>0$ is a residual floor (clamp) on $d_j$ and $d_0$, avoiding division by a near-zero denominator. The stop-gradient on $w_{\mathrm{sim}}$ makes $d_j$ and $d_0$ act only as a scalar loss reweighting: trajectories whose surrogate connectors better match the cached rollout receive larger training weight, while the gradient-routing graph remains the one specified by the active set, integration weights, and bridge scale. For connector-based methods without an interior bridge, the same rule reduces to endpoint-deviation weighting with the $d_j$ term omitted. Unless otherwise specified, we use loss scale 1.0, AdamW with learning rate $1\times10^{-5}$, weight decay $10^{-4}$, no learning-rate warmup, maximum gradient norm 1.0, EMA decay 0.995, and random seed 42. Online rollouts use 25 sampling steps at $512\times512$ resolution. We use classifier-free guidance scale 3.5 for \texttt{SD3.5-M} and 4.0 for the FLUX backbones. We train \texttt{SD3.5-M} for 250 iterations and both \texttt{FLUX.1-dev} and \texttt{FLUX.2-Klein-base} for 300 iterations. We enable gradient checkpointing, full-shard FSDP, TF32, EMA, and bf16 mixed precision. For non-CFG-distilled models, we detach the negative branch in classifier-free guidance.
\input{A-tab/method_hyperparameters}
\begin{table}[h]
\centering
\caption{Backbone-specific training configuration. All runs use HPSv2.1 as the training reward and 25-step online rollouts at $512\times512$ resolution.}
\label{tab:app_training_config}
\begin{tabular}{lccccccc}
\toprule
Backbone & GPUs & Iter. & CFG & Batch & LR & WD & EMA \\
\midrule
\texttt{SD3.5-M} & $8\times$H20 & 250 & 3.5 & 64 & $1\times10^{-5}$ & $10^{-4}$ & 0.995 \\
\texttt{FLUX.1-dev} & $16\times$H20 & 300 & 4.0 & 64 & $1\times10^{-5}$ & $10^{-4}$ & 0.995 \\
\texttt{FLUX.2-Klein-base} & $16\times$H20 & 300 & 4.0 & 64 & $1\times10^{-5}$ & $10^{-4}$ & 0.995 \\
\bottomrule
\end{tabular}
\end{table}

\noindent\textbf{Trajectory sampling.}
The choice of leap indices has a direct effect on optimization stability. Uniformly sampling two indices can place the two endpoints too close to each other or too close to the noisy end of the trajectory, which increases the variance of the reward gradient. To make the training signal more stable, we use a Dirichlet sampler to partition the valid reverse-index range into three segments. Let $N=25$ be the rollout length, let $r_{\mathrm{min}}$ and $r_{\mathrm{max}}$ denote the minimum and maximum reverse indices, and define $L=r_{\mathrm{max}}-r_{\mathrm{min}}$. We sample
\begin{equation}
    (p_a,p_b,p_c) \sim \mathrm{Dir}(\alpha_a,\alpha_b,\alpha_c),
    \qquad (\alpha_a,\alpha_b,\alpha_c)=(2.5,6.0,2.0),
\end{equation}
and convert the proportions into integer segment lengths
\begin{equation}
    (\ell_a,\ell_b,\ell_c) = \mathrm{Round}\bigl(L(p_a,p_b,p_c)\bigr),
    \qquad \ell_a+
    \ell_b+
    \ell_c=L,
\end{equation}
with a small correction to preserve the total length. The two reverse-time leap indices are then
\begin{equation}
    k_{\mathrm{rev}} = r_{\mathrm{min}} + \ell_a,
    \qquad
    j_{\mathrm{rev}} = r_{\mathrm{min}} + \ell_a + \ell_b,
\end{equation}
and the corresponding forward-time indices are
\begin{equation}
    k = N - k_{\mathrm{rev}},
    \qquad
    j = N - j_{\mathrm{rev}},
    \qquad 0 < j < k \le N,
\end{equation}
which is the strict two-segment case used by connector-style leap methods. The broader main-text convention is $N\ge k>j\ge0$, where $j=0$ denotes a single-segment surrogate without an interior bridge. Here, $\ell_a$ controls the location of $k$ (the noisy-side endpoint for leap-style methods), $\ell_b$ controls the gap between $k$ and $j$, and $\ell_c$ controls the location of $j$ (the clean-side split). For \ourscomposeleap{}, only the split $j$ is used and the noisy-side endpoint is fixed to $N$, consistent with its full Euler reconstruction. Unless otherwise stated, the nested-gradient scale is 0.3.

For \texttt{SD3.5-M} and \texttt{FLUX.2-Klein-base}, we additionally cap $k_{\mathrm{rev}}$ at $K_{\max}=10$ to avoid unstable placements near the noisy end. When $k_{\mathrm{rev}}>K_{\max}$, we truncate the first segment and move the overflow to the last segment:
\begin{equation}
    \ell_a' = \min(\ell_a, K_{\max}-r_{\mathrm{min}}),
    \qquad
    \ell_b' = \ell_b,
    \qquad
    \ell_c' = \ell_c + (\ell_a-\ell_a').
\end{equation}
This keeps the sampled gap $\ell_b$ unchanged while enforcing $k_{\mathrm{rev}}\leq K_{\max}$, which further stabilizes training.

\noindent\textbf{Surrogate supports and reconstruction.}
All our variants first perform a no-gradient rollout to cache trajectory states and velocities, and then re-forward only selected active velocities with gradients. For \oursleap, an interval from $s$ to $t$ (with $s>t$, so $\sigma_s>\sigma_t$) is approximated by a Lagrange connector with per-leap support set $\gS$,
\begin{equation}
    \hat\vx_t = \vx_s - \sum_{i\in\gS} w_i \tilde\vv_i,
    \qquad
    w_i = \int_{\sigma_t}^{\sigma_s}
    \prod_{q\in\gS, q\neq i}
    \frac{\sigma-\sigma_q}{\sigma_i-\sigma_q}\,d\sigma ,
\end{equation}
where the weights are non-negative for the small $M=|\gS|$ we use. For \ourscompose{}, the clean endpoint is reconstructed with Euler weights,
\begin{equation}
    \vx_0 = \vx_N - \sum_{i=1}^{N} h_i \tilde\vv_i,
    \qquad
    h_i = \sigma_i-\sigma_{i-1}>0 .
\end{equation}
For \ourscomposeleap{}, the same Euler reconstruction is applied separately on $(N\to j)$ and $(j\to0)$. Active steps in \ourscompose{} and \ourscomposeleap{} are sampled with a late-biased distribution $\Pr(i\in\A) \propto (N-i+1)^\beta$. Since only a subset $\A$ of velocities carries gradients, we use one shared gradient-rescaling rule for both high-order connectors and Euler reconstructions:
\begin{equation}
    g = 1 + \eta\left(
    \frac{\sum_{i\in\gS} |w_i|}{\sum_{i\in\A} |w_i|} - 1
    \right),
\end{equation}
where $w_i$ denotes either Lagrange weights or Euler weights $h_i$, and $\gS\supseteq\A$ is the full support or reconstruction range (the per-leap support set for \oursleap; $\{1,\ldots, N\}$ for \ourscompose{}; the per-segment Euler range $\{j+1,\ldots, N\}$ or $\{1,\ldots,j\}$ for \ourscomposeleap{}). We apply this factor to the re-forwarded active velocity (the regular active branch of \cref{eq:surrogate_v}, denoted $\vv_i$ below) via a stop-gradient identity,
\begin{equation}
    \tilde\vv_i = \sg(\vv_i) + g\bigl(\vv_i-\sg(\vv_i)\bigr),
    \qquad i\in\A,
\end{equation}
So the forward trajectory is unchanged while the active velocities receive a larger effective gradient. The concrete method-specific hyperparameters are summarized in \cref{tab:app_method_hyperparams}.

\noindent\textbf{FLUX.2-specific implementation.}
For \texttt{FLUX.2-Klein-base}, we precompute Qwen3 text embeddings using a maximum sequence length of 512 and concatenate hidden states from layers 9, 18, and 27 as the text-conditioning representation.

\noindent\textbf{Evaluation.}
We evaluate HPDv2 on the 400-prompt test split with 50-step sampling and report average scores and gains over the base model. GenEval results follow the official evaluator and report task accuracies and the overall score.

\noindent\textbf{Evaluation dynamics.}
In addition to final scores, \cref{fig:app_flux_eval_curves} reports PickScore, ImageReward, and HPSv2.1 over training for all three backbones, showing whether gains persist across checkpoints.

\input{A-fig/flux_eval_curves}
\FloatBarrier

%% file: A-tab/method_hyperparameters.tex
\begin{table}[t]
\centering
\caption{
Method-specific hyperparameters used for baselines and our variants.
Common optimization, reward, rollout, and precision settings are reported in
\cref{tab:app_training_config}.
Here, $K$ denotes the number of active steps,
$\beta$ is the late-bias exponent for active-step sampling,
$\eta$ is the gradient-rescale strength,
$\alpha$ is the nested-gradient scale,
$g_s$ is the gradient-support scale,
$\tau$ is the residual floor, and $n$ is the number of noised samples in DRaFT-LV.
}
\label{tab:app_method_hyperparams}
\setlength{\tabcolsep}{3pt}
\renewcommand{\arraystretch}{1.1}
\begin{tabular}{llp{0.54\linewidth}}
\toprule
Method & Model & Hyperparameters \\
\midrule

ReFL
& All
& Last-step window $N_{\mathrm{last}}=11$ \\

DRaFT-LV
& All
& Noised samples $n=2$ \\

DRTune
& All
& Active train steps $K=3$; early-stop ratio $0.4$ \\

LeapAlign
& All
& $\alpha=0.3$; trajectory weighting $\tau=0.1$ \\

\midrule

\multirow{4}{*}{\oursleap{}}
& Shared
& Dirichlet sampler $(2.5,6.0,2.0)$; $g_s = 0.25$; $\alpha=0.1$ \\
& \texttt{SD3.5-M}
& $\tau=0.2,\; k_{\rm rev}\le10,\; \eta=0$ \\
& \texttt{FLUX.1-dev}
& $\tau=0.2,\; \eta=0.5$ \\
& \texttt{FLUX.2-Klein-base}
& $\tau=0.2,\; k_{\rm rev}\le10,\; \eta=0$ \\

\midrule

\multirow{4}{*}{\ourscompose{}}
& Shared
& Active steps sampled with $\Pr(i\in\A)\propto(N-i+1)^\beta$ \\
& \texttt{SD3.5-M}
& $K=4,\; \beta=2,\; \eta=0$ \\
& \texttt{FLUX.1-dev}
& $K=3,\; \beta=4,\; \eta=1$ \\
& \texttt{FLUX.2-Klein-base}
& $K=4,\; \beta=2,\; \eta=0$ \\

\midrule

\multirow{4}{*}{\ourscomposeleap{}}
& Shared
& Active steps sampled with $\Pr(i\in\A)\propto(N-i+1)^\beta$, $\beta=4$ \\
& \texttt{SD3.5-M}
& $K=5,\; \eta=0.5,\; \alpha=0.3$ \\
& \texttt{FLUX.1-dev}
& $K=5,\; \eta=1,\; \alpha=0.1$ \\
& \texttt{FLUX.2-Klein-base}
& $K=3,\; \eta=0.5,\; \alpha=1.0$ \\

\bottomrule
\end{tabular}
\end{table}

%% file: A-fig/flux_eval_curves.tex
\begin{figure}[!htbp]
\centering
\begin{minipage}{1.0\textwidth}
    \centering
    \includegraphics[width=\textwidth]{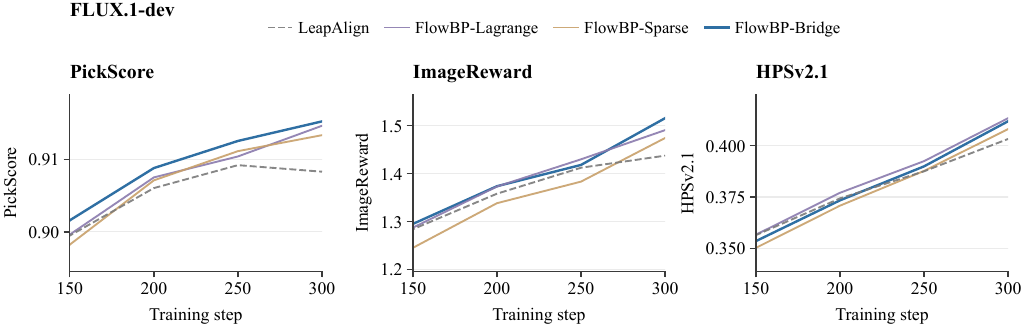}
\end{minipage}

\vspace{0.25em}

\begin{minipage}{1.0\textwidth}
    \centering
    \includegraphics[width=\textwidth]{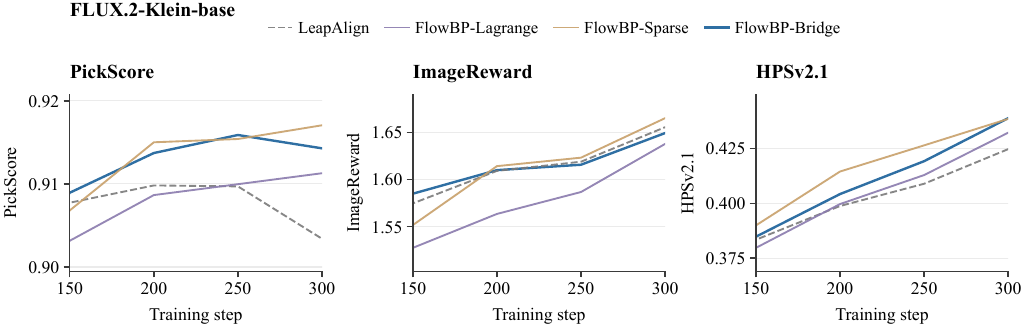}
\end{minipage}

\vspace{0.25em}

\begin{minipage}{1.0\textwidth}
    \centering
    \includegraphics[width=\textwidth]{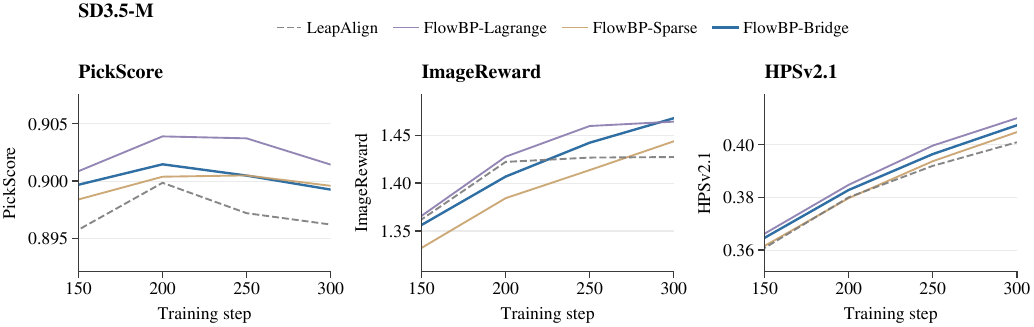}
\end{minipage}

\caption{Evaluation dynamics on the HPDv2 test split for \texttt{FLUX.1-dev}, \texttt{FLUX.2-Klein-base}, and \texttt{SD3.5-M}. Each row reports PickScore, ImageReward, and HPSv2.1 over training; PickScore is normalized by $26$.}
\label{fig:app_flux_eval_curves}
\end{figure}

%% file: A-sec/03_algorithm_templates.tex
\section{Algorithmic Templates for Reward-Gradient Updates}
\label{appsec:algorithm_templates}

This appendix specifies the reward-gradient update graphs used by the methods discussed in the main text. Each template identifies the reward input, the active velocity evaluations, and the surrogate path through which gradients are propagated.

We present two templates. Posterior-mean reward updates query the reward on a short clean estimate, whereas cached-rollout surrogate updates first sample a full trajectory without gradients and then construct a sparse backward graph on top of the cached states. Together, the two templates make the gradient-routing differences between the compared methods explicit.

\subsection{Posterior-Mean Reward Updates}
\label{appsec:posterior_mean_template}

ReFL, DRTune, and DRaFT-LV can be written as instances of a common posterior-mean update. They differ in the stop index, the set of velocity evaluations that remain active for gradients, and the number of clean targets used to evaluate the reward. The shared template is given in \cref{alg:baseline_unified}.

\input{A-tab/baseline_algorithm}

\subsection{Cached-Rollout Surrogate Updates}
\label{appsec:surrogate_template}

LeapAlign and the proposed FlowBP variants use a different graph construction. They first generate and cache a complete rollout without storing activations, and then expose only selected velocity evaluations to autograd. Connector-based variants attach sparse predictions to cached states with straight-through connectors, whereas reconstruction-based variants replay Euler updates with cached inactive velocities. \Cref{alg:surrogate_unified} summarizes this family of update graphs.

\input{A-tab/surrogate_algorithm}

%% file: A-tab/baseline_algorithm.tex
\makeatletter
\@ifundefined{c@algorithm}{\newcounter{algorithm}}{}
\makeatother
\renewcommand{\thealgorithm}{\arabic{algorithm}}
\providecommand{\sg}{\operatorname{sg}}

\definecolor{stagec}{RGB}{45,92,170}
\definecolor{reflc}{RGB}{31,119,180}
\definecolor{drtunec}{RGB}{214,39,40}
\definecolor{draftc}{RGB}{148,103,189}
\definecolor{leapc}{RGB}{44,160,44}
\definecolor{highc}{RGB}{230,126,34}
\definecolor{composec}{RGB}{23,162,184}
\definecolor{splitc}{RGB}{127,79,36}
\newcommand{\refltag}{\textcolor{reflc}{\textnormal{ReFL}}}
\newcommand{\drtunetag}{\textcolor{drtunec}{\textnormal{DRTune}}}
\newcommand{\drafttag}{\textcolor{draftc}{\textnormal{DRaFT-LV}}}
\newcommand{\leaptag}{\textcolor{leapc}{\textnormal{LeapAlign}}}
\newcommand{\hightag}{\textcolor{highc}{\textnormal{\oursleap{}}}}
\newcommand{\composetag}{\textcolor{composec}{\textnormal{\ourscompose{}}}}
\newcommand{\splittag}{\textcolor{splitc}{\textnormal{\ourscomposeleap{}}}}
\newcommand{\algstage}[1]{\item[] \textcolor{stagec}{\textbf{#1}}}

\newcommand{\algtitle}[2]{%
  \refstepcounter{algorithm}%
  \noindent\textbf{Algorithm~\thealgorithm: #1}\label{#2}\par\vspace{4pt}%
}
\newenvironment{algblock}[2]{%
  \begin{center}
  \begin{minipage}{1.00\textwidth}
  \small
  \setlength{\abovedisplayskip}{2pt}
  \setlength{\belowdisplayskip}{2pt}
  \setlength{\abovedisplayshortskip}{2pt}
  \setlength{\belowdisplayshortskip}{2pt}
  \hrule\vspace{3pt}
  \algtitle{#1}{#2}
  \vspace{0pt}\hrule\vspace{3pt}
}{%
  \vspace{2pt}\hrule
  \end{minipage}
  \end{center}
}
\newenvironment{algsteps}{%
  \begin{enumerate}[label=\arabic*:,leftmargin=2.35em,labelsep=0.5em,itemsep=2pt,parsep=0.5pt,topsep=2pt]
}{%
  \end{enumerate}
}
\newcommand{\algindent}{\hspace*{1.25em}}
\newcommand{\alginout}[2]{\par\noindent\hangindent=4.8em\hangafter=1\textbf{#1:}\ #2\par}

\begin{algblock}{Unified Posterior-Mean Reward Updates}{alg:baseline_unified}
\alginout{Input}{Velocity model $\vv_\theta$; reward function $r(\cdot,\cdot)$; prompt distribution $p_c$; hinge threshold $\lambda$; noise schedule $\{\sigma_i\}_{i=0}^{N}$; method flag $m\in\{\refltag,\drtunetag,\drafttag\}$.}
\alginout{Output}{Updated model parameters $\theta$.}
\begin{algsteps}
\algstage{Unified rollout graph}
\item Sample $\vc\sim p_c$, $\vx_N\sim\mathcal{N}(\mathbf{0},\mathbf{I})$.
\item Choose a stop index $e$, active set $\A$, and clean-prediction index set $\gQ$:
\item[] \algindent \refltag: $(e,\A,\gQ)=(\tau,\{\tau\},\{0\})$, where $\tau \in\gT_{\mathrm{tail}}$.
\item[] \algindent \drtunetag: $(e,\A,\gQ)=(t_{\min},\,t_{\mathrm{train}}\cup\{t_{\min}\},\,\{0\})$.
\item[] \algindent \drafttag: $(e,\A,\gQ)=(1,\{1\},\{0,\ldots,n\})$.
\item For all executed steps, use
\[
\tilde\vv_i=
\begin{cases}
 \vv_\theta(\sg(\vx_i),\sigma_i,\vc), & i\in\A,\\
 \sg\!\bigl(\vv_\theta(\sg(\vx_i),\sigma_i,\vc)\bigr), & i\notin\A,
\end{cases}
\qquad
\vx_{i-1}=\vx_i-h_i\,\tilde\vv_i,
\qquad h_i=\sigma_i-\sigma_{i-1}>0 .
\]
\algstage{Unified reward targets}
\item Define the shared posterior-mean target
\[
\vz_0=\vx_e-\sigma_e\,\tilde\vv_e .
\]
\item For \drafttag{} only, keep $\vz_0$ as the first reward target ($q=0$), corresponding to the direct update before re-noising. Then construct $n$ additional low-variance targets by re-noising $\vz_0$:
\[
\vx_1^{(q)}=(1-\sigma_1)\sg(\vz_0)+\sigma_1\vepsilon_q,
\qquad
\vz_q=\vx_1^{(q)}-\sigma_1\,\vv_\theta(\sg(\vx_1^{(q)}),\sigma_1,\vc),
\quad q=1,\ldots,n.
\]
\algstage{Reward update}
\item Compute $\mathcal{L}=|\gQ|^{-1}\sum_{q\in\gQ}\operatorname{ReLU}\bigl(\lambda-r(\vz_q,\vc)\bigr)$, where the single-target cases use $\gQ=\{0\}$ and therefore only the $\vz_0$ term.
\item Update $\theta\leftarrow\theta-\eta\nabla_\theta\mathcal{L}$.
\item \textbf{return} $\theta$.
\end{algsteps}
\end{algblock}

%% file: A-tab/surrogate_algorithm.tex
\begin{algblock}{Unified Surrogate-Trajectory Gradient Methods}{alg:surrogate_unified}
\alginout{Input}{Velocity model $\vv_\theta$; scalar reward $r(\vx_0,\vc)$; prompt distribution $p_c$; hinge threshold $\lambda$; noise schedule $\{\sigma_i\}_{i=0}^{N}$; method flag $m\in\{\leaptag,\hightag,\composetag,\splittag\}$.\par}
\alginout{Output}{Updated model parameters $\theta$.}
\begin{algsteps}
\algstage{Unified cached rollout}
\item Sample $\vc\sim p_c$, $\vx_N\sim\mathcal{N}(\mathbf{0},\mathbf{I})$.
\item Run a no-gradient rollout and cache $\{\vx_i,\vv_i,\sigma_i\}_{i=0}^{N}$.
\item Choose indices, active sets $\A=\Apre\cup\Apost$, and weights $\{w_i\}$:
\item[] \algindent \leaptag: choose $0<j<k\le N$, $\Apre=\{k\}$, $\Apost=\{j\}$, $w_k=\sigma_k-\sigma_j$, $w_j=\sigma_j$.
\item[] \algindent \hightag: choose $0<j<k\le N$, Lagrange supports $\mathcal{S}_{\mathrm{pre}}$ and $\mathcal{S}_{\mathrm{post}}$ anchored at $k$ and $j$, active sets $\Apre\subseteq\mathcal{S}_{\mathrm{pre}}$, $\Apost\subseteq\mathcal{S}_{\mathrm{post}}$ with $|\Apre\cup\Apost|\le M$, and $w_i=w_i^{\mathrm{L}}$.
\item[] \algindent \composetag: $j=0$, $\Apost=\emptyset$, $\A=\Apre\subseteq\{1,\ldots,N\}$, $w_i=h_i$.
\item[] \algindent \splittag: choose $j\in\{1,\ldots,N-1\}$, $\Apre\subseteq\{j+1,\ldots,N\}$, $\Apost\subseteq\{1,\ldots,j\}$ with $j\in\Apost$, $w_i=h_i$.
\item Define active/inactive velocities (\cref{eq:surrogate_v})
\algstage{Unified surrogate targets}
\item For \hightag{}, let $\bar\vv_i$ denote the Lagrange-support velocity: it follows \cref{eq:surrogate_v} when $i\in\A$ (including the bridge case at $i=j$), and uses the detached cached velocity $\sg(\vv_i)$ when $i\notin\A$.
\item For connector-based bridged methods, attach the pre-segment surrogate to the cached bridge state:
\[
\hat\vx_j=
\begin{cases}
\vx_k-w_k\,\tilde\vv_k, & \leaptag,\\[2pt]
\vx_k-\sum_{i\in\mathcal{S}_{\mathrm{pre}}}w_i^{\mathrm{L}}\,\bar\vv_i, & \hightag,
\end{cases}
\qquad
\vx_j=\hat\vx_j+\sg(\vx_j-\hat\vx_j).
\]
\item For \splittag{}, reconstruct the bridge state by Euler reconstruction
\[
\vx_j=\vx_N-\sum_{i=j+1}^{N}h_i\,\tilde\vv_i.
\]
\item Define the clean endpoint
\[
\vx_0=
\begin{cases}
\hat\vx_0+\sg(\vx_0-\hat\vx_0),\quad
\hat\vx_0=\vx_j-w_j\,\tilde\vv_j, & \leaptag,\\[2pt]
\hat\vx_0+\sg(\vx_0-\hat\vx_0),\quad
\hat\vx_0=\vx_j-\sum_{i\in\mathcal{S}_{\mathrm{post}}}w_i^{\mathrm{L}}\,\bar\vv_i, & \hightag,\\[2pt]
\vx_N-\sum_{i=1}^{N}h_i\,\tilde\vv_i, & \composetag,\\[2pt]
\vx_j-\sum_{i=1}^{j}h_i\,\tilde\vv_i, & \splittag.
\end{cases}
\]
\algstage{Reward update}
\item Compute $\mathcal{L}=\operatorname{ReLU}(\lambda-r(\vx_0,\vc))$.
\item Update $\theta\leftarrow\theta-\eta\nabla_\theta\mathcal{L}$ through active calls only.
\item \textbf{return} $\theta$.
\end{algsteps}
\end{algblock}

%% file: A-sec/03_qualitative_results.tex
\section{Additional Qualitative Results}
\label{appsec:qualitative_results}

We provide additional qualitative examples generated from HPDv2 test prompts for \texttt{SD3.5-M}, \texttt{FLUX.1-dev}, and \texttt{FLUX.2-Klein-base} in \cref{fig:app_sd35_qualitative_set1,fig:app_flux1_qualitative_set1,fig:app_flux2_qualitative_set1}, respectively.
\input{A-fig/sd35_qualitative_comparison_set1}
\input{A-fig/Flux1_qualitative_comparison_set1}
\input{A-fig/flux2_qualitative_comparison_set1}

%% file: A-fig/sd35_qualitative_comparison_set1.tex
\begin{figure}[h]
\centering
\includegraphics[width=\textwidth,height=0.85\textheight,keepaspectratio]{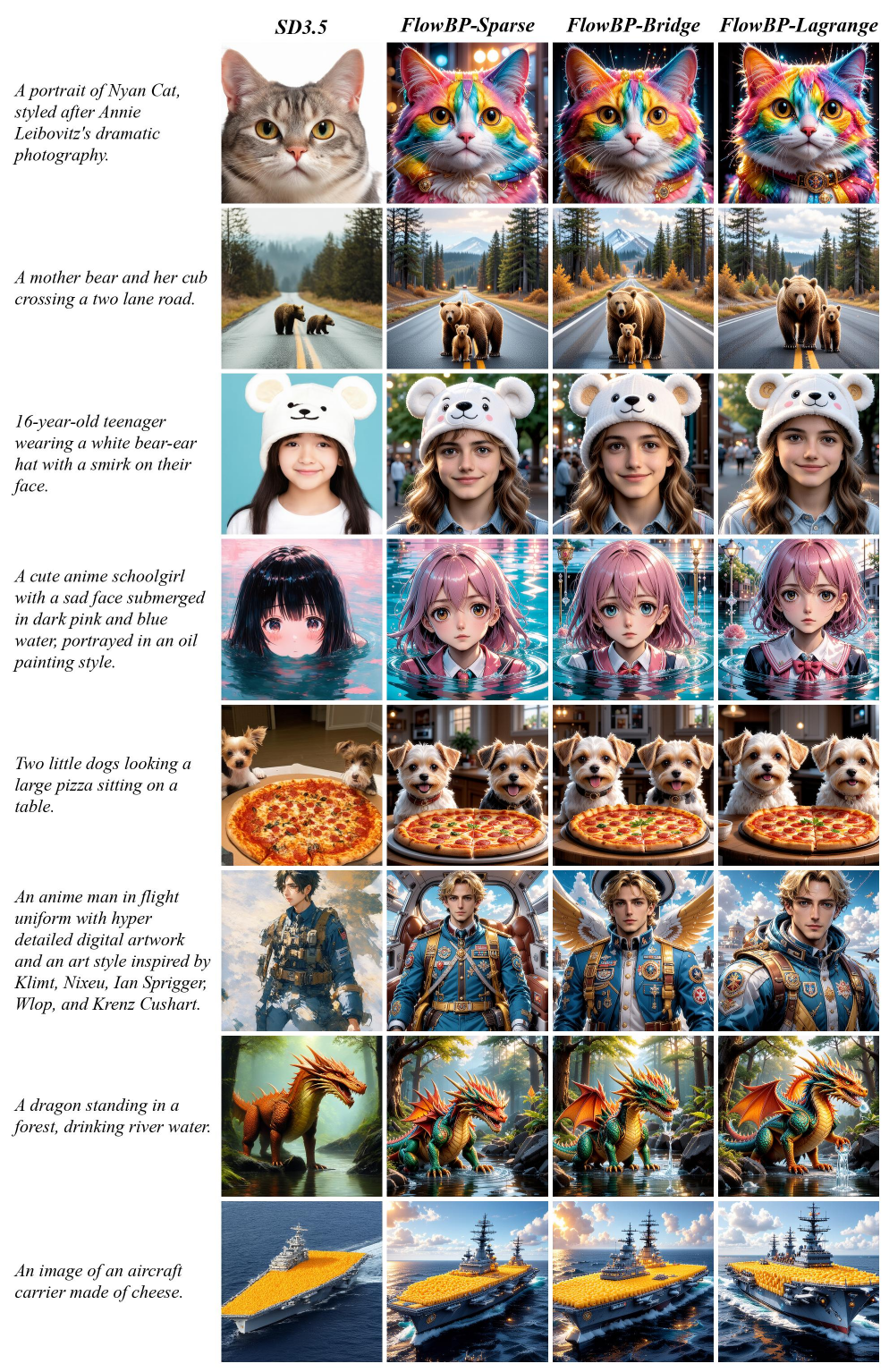}
\caption{Additional qualitative results on \texttt{SD3.5-M} using prompts from the HPDv2 test split.}
\label{fig:app_sd35_qualitative_set1}
\end{figure}

%% file: A-fig/Flux1_qualitative_comparison_set1.tex
\begin{figure}[t]
\centering
\includegraphics[width=\textwidth,height=0.95\textheight,keepaspectratio]{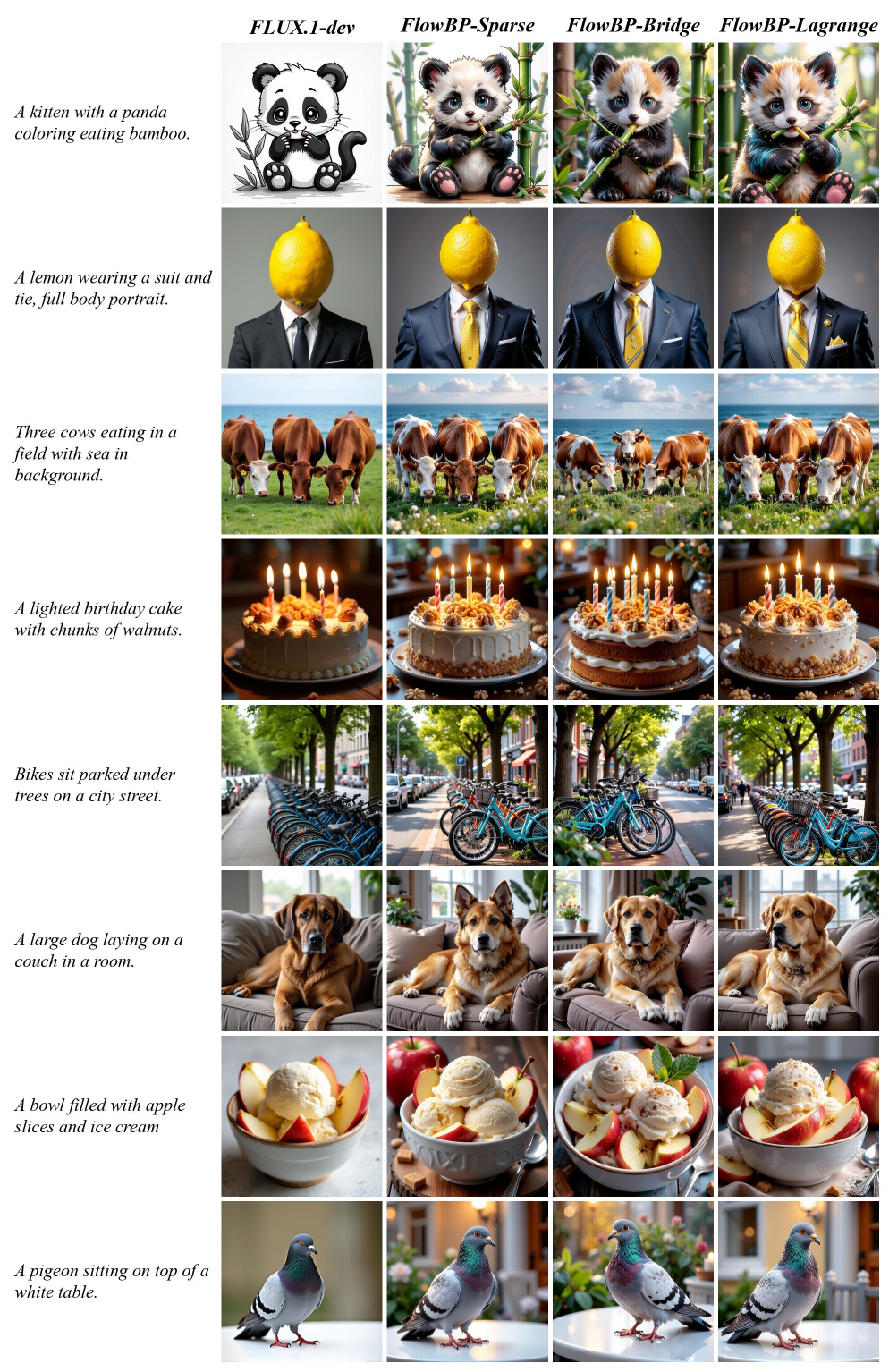}
\caption{Additional qualitative results on \texttt{FLUX.1-dev} using prompts from the HPDv2 test split.}
\label{fig:app_flux1_qualitative_set1}
\end{figure}

%% file: A-fig/flux2_qualitative_comparison_set1.tex
\begin{figure}[t]
\centering
\includegraphics[width=\textwidth,height=0.95\textheight,keepaspectratio]{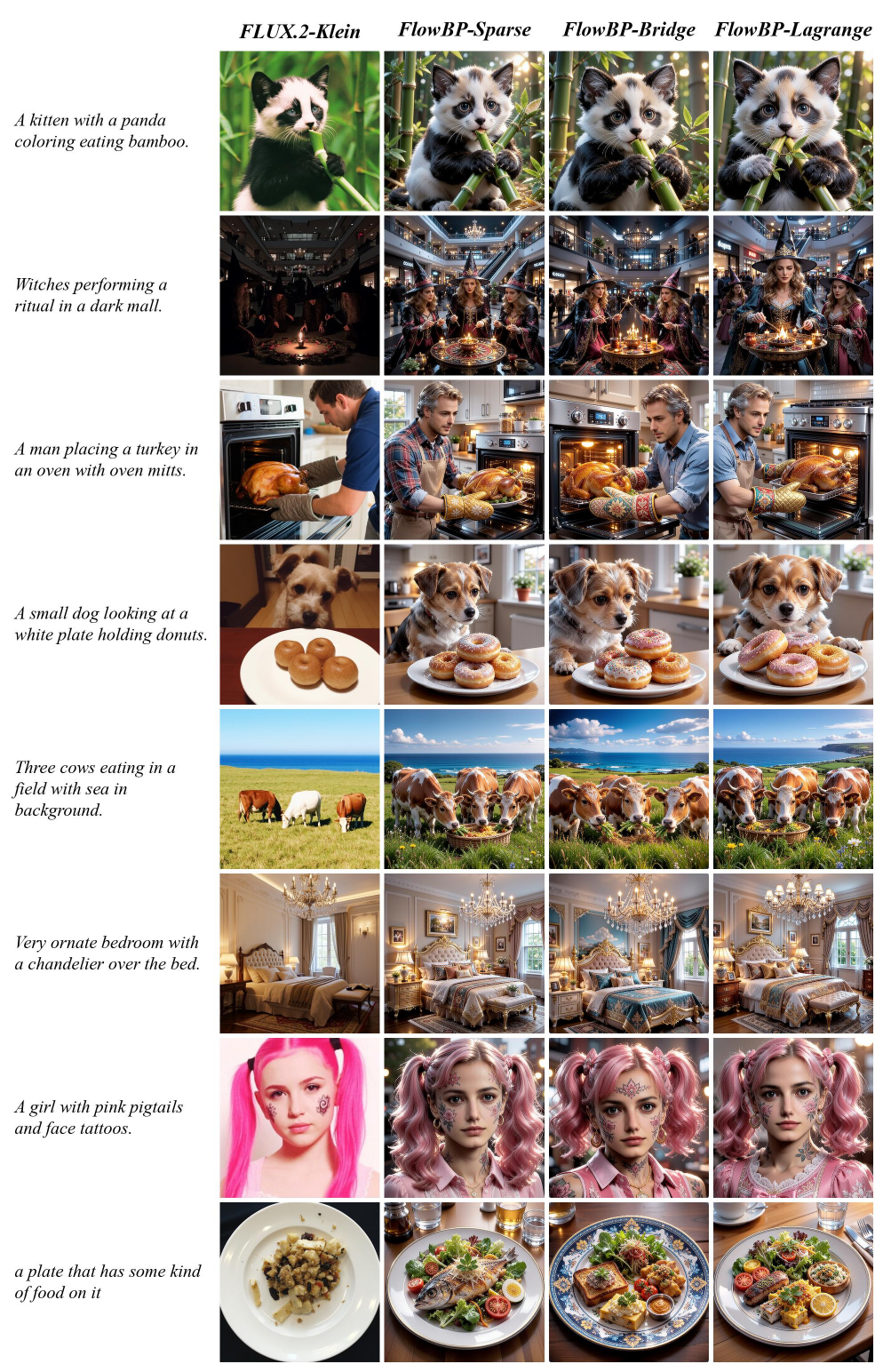}
\caption{Additional qualitative results on \texttt{FLUX.2-Klein-base} using prompts from the HPDv2 test split.}
\label{fig:app_flux2_qualitative_set1}
\end{figure}